\documentclass{article} 
\usepackage[preprint]{colm2026_conference}

\usepackage{microtype}
\usepackage{hyperref}
\usepackage{url}
\usepackage{booktabs}
\usepackage{enumitem}
\newcommand{\name}{{\textsc{M2-Verify}}}
\usepackage{multirow}

\usepackage{graphicx}
\usepackage{caption}
\usepackage[most]{tcolorbox}
\usepackage{xcolor}
\usepackage[table,dvipsnames]{xcolor}

\usepackage{fontawesome5}
\usepackage{graphicx}

\definecolor{lightbluebox}{RGB}{220,235,255}
\definecolor{weakgreen}{RGB}{220, 255, 220}
\definecolor{weakred}{RGB}{255, 220, 220}

\usepackage{arydshln}
\usepackage{CJKutf8}
\usepackage{amssymb}

\usepackage{lineno}

\definecolor{darkblue}{rgb}{0, 0, 0.5}
\hypersetup{colorlinks=true, citecolor=darkblue, linkcolor=darkblue, urlcolor=darkblue}

\title{M2-Verify: A Large-Scale Multidomain Benchmark for Checking Multimodal Claim Consistency}

\author{%
  \textbf{Abolfazl Ansari}\textsuperscript{1}, 
  \textbf{Delvin Ce Zhang}\textsuperscript{2}, 
  \textbf{Zhuoyang Zou}\textsuperscript{1}, 
  \textbf{Wenpeng Yin}\textsuperscript{1}, 
  \textbf{Dongwon Lee}\textsuperscript{1} \\
  \textsuperscript{1}The Pennsylvania State University, USA \\
  \textsuperscript{2}University of Sheffield, UK \\
  \texttt{\{aja7154, zhuoyangzou, wenpeng, dul13\}@psu.edu} \\
  \texttt{delvincezhang@gmail.com}
}

\usepackage{xcolor}

\begin{document}

\ifcolmsubmission
\linenumbers
\fi

\maketitle

\begin{abstract}

Evaluating scientific arguments requires assessing the strict consistency between a claim and its underlying multimodal evidence. However, existing benchmarks lack the scale, domain diversity, and visual complexity needed to evaluate this alignment realistically. To address this gap, we introduce {\name}, a large-scale multimodal dataset for checking scientific claim consistency. Sourced from PubMed and arXiv, {\name} provides over 469K instances across 16 domains, rigorously validated through expert audits. Extensive baseline experiments show that state-of-the-art models struggle to maintain robust consistency. While top models achieve up to 85.8\% Micro-F1 on low-complexity medical perturbations, performance drops to 61.6\% on high-complexity challenges like anatomical shifts. Furthermore, expert evaluations expose hallucinations when models generate scientific explanations for their alignment decisions. Finally, we demonstrate our dataset's utility and provide comprehensive usage guidelines.

\footnote{Hugging Face: \href{https://huggingface.co/datasets/AbolfazlAnsari/M2-Verify-Med}{M2-Verify-Med}, \href{https://huggingface.co/datasets/AbolfazlAnsari/M2-Verify-Gen}{M2-Verify-Gen}. GitHub: \href{https://github.com/M2-Verify/M2-Verify}{M2-Verify/M2-Verify}.}

\end{abstract}
\section{Introduction}


Scientific claim consistency checking evaluates whether a stated claim is strictly supported by the specific multimodal evidence provided within a closed document. As scholarly literature expands, the need for automated systems capable of this rigorous consistency checking has become increasingly important~\citep{vladika-matthes-2023-scientific}. An effective consistency checking system must be able to evaluate claims grounded across different modalities, reflecting the inherently multimodal nature of scientific communication~\citep{Sciver2025acl}. Moreover, the structure and representation of scientific figures vary widely across domains, making multimodal consistency checking even more challenging.

Authors rely on diverse visual elements, ranging from quantitative charts to medical scans, coupled with explanatory text to support findings~\citep{Sciver2025acl}. As a result, evaluating whether conclusions are fully consistent with this multimodal evidence requires cross-modal reasoning. This demand has become more critical with the advancement of generative AI~\citep{ramesh2021zeroshottexttoimagegeneration, rombach2022highresolutionimagesynthesislatent}. As these tools are increasingly used to generate or enhance scientific figures, there is a risk that claims rely on visuals containing hallucinations or artifacts, directly threatening scientific integrity~\citep{ethicsofimage}. This underscores the critical need for claim consistency evaluation.

\begin{table*}[t]
\centering
\footnotesize
\renewcommand{\arraystretch}{1.3} 
\setlength{\tabcolsep}{3.5pt}

\resizebox{\textwidth}{!}{%

\begin{tabular}{@{} l >{\centering\arraybackslash}m{3.4cm} >{\centering\arraybackslash}m{3.4cm} c c c c @{}} 
\toprule
\textbf{Dataset} & \textbf{Domain} & \textbf{Source} & \textbf{Expl.} & \textbf{\# Figs} & \textbf{\# Dom} & \textbf{Size} \\
\midrule

\multicolumn{7}{c}{\textbf{Text-only Claim Verification (General Domain)}} \\
\midrule
FEVER \citep{thorne-etal-2018-fever} & General & Wikipedia & No & 0 & 0 & \cellcolor{weakgreen}185K \\
PubHealth \citep{PubHealth2020} & \cellcolor{weakgreen}Public Health & Fact-check sites, News & \cellcolor{weakgreen}Yes & 0 & 0 & 11.8K \\

\midrule
\multicolumn{7}{c}{\textbf{Text-only Claim Verification (Scientific Domain)}} \\
\midrule
SciFact \citep{Wadden2020a} & \cellcolor{weakgreen}Biomedical & \cellcolor{weakgreen}Semantic Scholar & \cellcolor{weakgreen}Yes & 0 & 1 & 1.4K \\
Climate-FEVER \citep{diggelmann2020climatefever} & \cellcolor{weakgreen}Climate Science & Wikipedia & \cellcolor{weakgreen}Yes & 0 & 1 & 1.5K \\
COVID-Fact \citep{COVIDFact2021} & \cellcolor{weakgreen}Biom. + Pub. Health & Sci. articles + Media (Web) & \cellcolor{weakgreen}Yes & 0 & 1 & 4.1K \\
CoVERT \citep{CoVERT2022} & \cellcolor{weakgreen}Biomedical (COVID-19) & Twitter + Web Evidence & \cellcolor{weakgreen}Yes & 0 & 1 & 300 \\

\midrule
\multicolumn{7}{c}{\textbf{Multimodal Claim Verification (General Domain)}} \\
\midrule
FEVEROUS \citep{Aly2021} & General & Wikipedia (Text + Tables)& No & 0 & 0 & \cellcolor{weakgreen}87K \\
MOCHEG \citep{MOCHEG} & General & PolitiFact, Snopes & \cellcolor{weakgreen}Yes & $\dagger$ & 0 & 15.6K \\
ChartCheck \citep{Akhtar2024ChartCheck} & General & Wikimedia Commons & \cellcolor{weakgreen}Yes & 1 & 0 & 10.5K \\

\midrule
\multicolumn{7}{c}{\textbf{Multimodal Claim Verification (Scientific Domain)}} \\
\midrule
MuSciClaims \citep{Lal2025MuSciClaims} & \cellcolor{weakgreen}Bio., Chem., Cell & \cellcolor{weakgreen}Nature, JACS, Cell & No & \cellcolor{weakgreen}4 & \cellcolor{weakgreen}3 & 1.5K \\
SciVer \citep{Sciver2025acl} & \cellcolor{weakgreen}Computer Science & \cellcolor{weakgreen}arXiv (CS) & \cellcolor{weakgreen}Yes & 2 & 1 & 3K \\

\midrule
\textbf{\name{} (Ours)} & \cellcolor{weakgreen}Medical + Gen. Science & \cellcolor{weakgreen}PubMed + arXiv & \cellcolor{weakgreen}Yes & \cellcolor{weakgreen}6 & \cellcolor{weakgreen}16 & \cellcolor{weakgreen}469K \\
\bottomrule
\end{tabular}
}
\vspace{-2mm}
\caption{
Comparison of claim verification datasets across domains. Datasets are grouped by modality and domain. Green indicates scientific coverage or largest instances. ${\dagger}$ MOCHEG contains heterogeneous visual evidence and does not define a fixed set of figure types.
}
\label{tab:all_datasets}
\vspace{-4mm}
\end{table*}

Text-only benchmarks provide a foundation for scientific consistency evaluation and claim verification, but remain confined to unimodal settings~\citep{Wadden2020a, Wang2023Covid}. This limitation overlooks the complementary nature of modalities. While textual descriptions provide context, granular, domain-specific evidence is encoded exclusively within the visual modality. This results in an informational asymmetry. Without cross-modal grounding, text-only systems lack the necessary evidence to verify claims. Moreover, visual evidence in scientific literature encodes domain-specific semantics that may not be interchangeable across fields. 

\begin{figure*}[t]
\centering

\includegraphics[height=0.4\textheight]{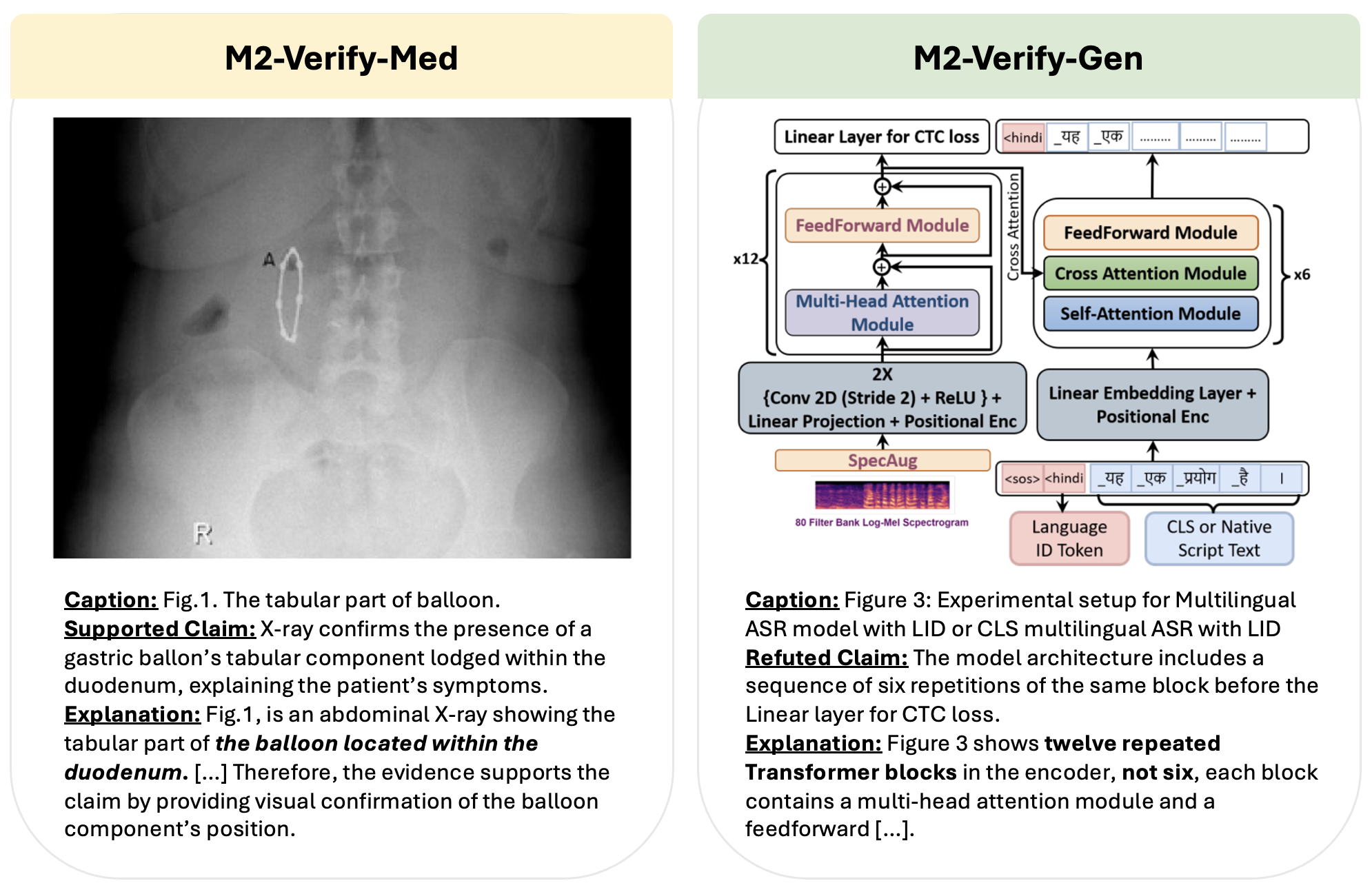}
\par\noindent

\vspace{-2mm}
\caption{Representative examples from {\name}-Med and {\name}-Gen exemplifying its multimodal diversity. Verifying gastric balloon location (left) and model architecture (right) requires joint reasoning across figures and captions.}\label{fig:data_samples}
\vspace{-5mm}
\end{figure*}

To address these gaps, we introduce \textbf{{\name}}, a large-scale multidomain benchmark for assessing multimodal consistency and supporting explainable verification. Covering 16 scientific domains, {\name} integrates over 469K instances of scientific figures, captions, and claims derived from PubMed and arXiv. The dataset captures a diverse array of visual evidence, including medical imaging, model architectures, and quantitative charts, each paired with a natural language explanation that grounds the verification decision in the multimodal evidence (Figure~\ref{fig:data_samples}).

To enable evaluation of multimodal scientific claim consistency, we introduce a calibrated pipeline rigorously validated by a three phase audit involving 92 domain experts. {\name} provides fine grained diagnostic utility through a taxonomy of medical perturbations and detailed natural language explanations to explicitly assess model faithfulness. Utilizing this benchmark, our evaluation of 12 leading vision language models reveals severe cross modal reasoning deficits. Top models exhibit a sharp performance drop from 85.8\% Micro F1 on simple visual shifts to 61.6\% on complex anatomical shifts, while also struggling significantly in abstract fields like mathematics and physics. Finally, ablation studies demonstrate that both figures and captions contribute independently, highlighting the strict necessity of integrated multimodal reasoning for scientific claim consistency. In summary, our core contributions are as follows.
\vspace{-3mm}
\paragraph{Contributions.} (1) We introduce {\name}, a 469K-instance benchmark across 16 disciplines, with two subsets: {\name}-Med for the medical domain with perturbations, and {\name}-Gen for general science fields from arXiv. (2) We construct a generation pipeline featuring domain specific perturbation taxonomies validated by a three phase audit from 92 experts. (3) We evaluate 12 vision language models exposing drops on complex visual shifts and hallucinations before demonstrating task learnability via supervised fine-tuning.

\vspace{-2mm}
\section{Related Work}

\vspace{-3mm}

\noindent \textbf{Text-only Claim Verification.} 
Foundational benchmarks in the general domain, such as FEVER \citep{thorne-etal-2018-fever} and PubHealth \citep{PubHealth2020}, grounded verification in Wikipedia and news articles. To address specialized science context, text-only scientific benchmarks were developed: SciFact \citep{Wadden2020a} and COVID-Fact \citep{COVIDFact2021} target biomedical claims, while Climate-FEVER \citep{diggelmann2020climatefever} and CoVERT \citep{CoVERT2022} focus on climate science and online medical misinformation. Furthermore, recent surveys \citep{dmonte2024claim} highlight the increasing reliance on LLMs for these tasks, alongside their expansion into automated peer review and weakness identification \citep{zou2026diagpaper}. However, while effective for textual consistency, existing benchmarks remain constrained to text-only evidence, ignoring the complex figures and charts essential to valid scientific communication.


\noindent \textbf{Multimodal Claim Verification.} 
Recent work incorporates non-textual evidence to address this limitation. In the general domain, FEVEROUS~\citep{Aly2021}, ChartCheck~\citep{Akhtar2024ChartCheck}, and MOCHEG~\citep{MOCHEG} integrate tables and open-domain images but lack the rigor of academic literature. Other studies have expanded this to multi-hop multimodal claim verification~\citep{wang2025piecing}. Conversely, emerging scientific multimodal datasets like MuSciClaims~\citep{Lal2025MuSciClaims} and SciVer~\citep{Sciver2025acl} utilize academic figures but remain severely constrained in overall scale, domain coverage, limited primarily to biology or computer science, and figure diversity. Consequently, there remains a critical need for a benchmark offering both high-granularity visual evidence and broad cross-disciplinary coverage. As summarized in Table~\ref{tab:all_datasets}, \name{} uniquely addresses this gap by combining large-scale coverage (469K instances), diverse scientific fields (16 domains), and explainability.

\vspace{-3mm}
\section{{\name} Dataset}
\begin{figure*}[t]
    \centering
    \includegraphics[width=\textwidth]{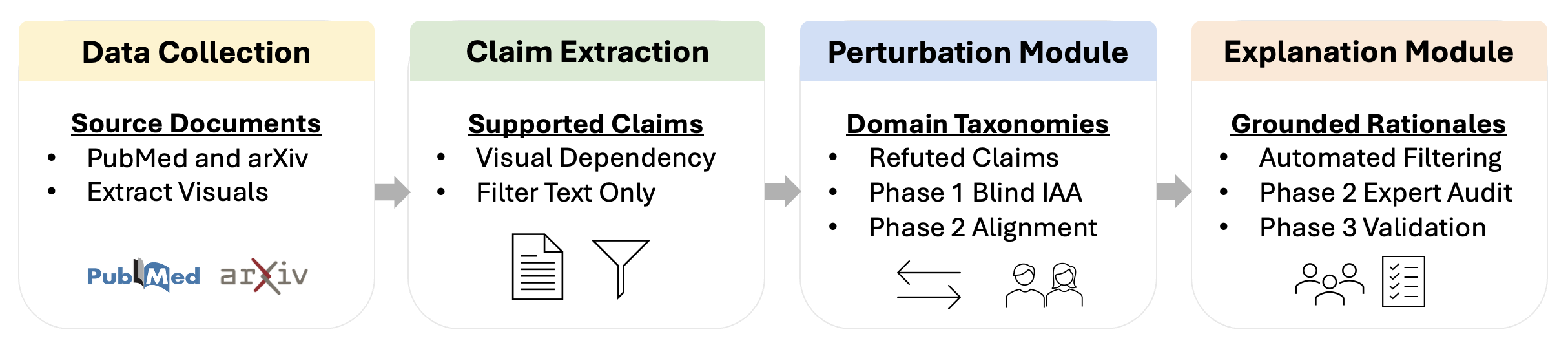}
\caption{Overview of the {\name} framework. The pipeline integrates automated claim extraction visual dependency filtering domain specific perturbations and grounded explanations validated by multi phase expert audit.}
    \label{fig:framework}
    \vspace{-6mm}
\end{figure*}

\vspace{-4mm}
Figure~\ref{fig:framework} provides an overview of the {\name} framework. It employs a framework consisting of four calibrated stages: (1) collecting figures and captions from scientific papers where authors explicitly draw conclusions from visual evidence; (2) extracting supported claims grounded in both figures and captions; (3) applying customized perturbations to generate refuted claims; and (4) crafting explanations to justify veracity labels by rationalizing the relevant visual and textual evidence. Expert evaluation is applied to ensure the quality of both claims and explanations, resulting in a reliable dataset.


\vspace{-3mm}
\subsection{Data Construction}
\vspace{-3mm}
\noindent\textbf{Data Collection.} To construct a multimodal scientific claim verification dataset spanning diverse domains, {\name} builds upon two existing open-source datasets: MedICaT \citep{subramanian-etal-2020-medicat} and SciMMIR \citep{wu-etal-2024-scimmir}. While these datasets provide large-scale collections of scientific figures and captions, they were primarily designed for figure-centric retrieval and alignment tasks rather than claim verification. As such, they lack the specific claim-evidence structures necessary for claim consistency task. Since all instances in these datasets are derived from published papers, claims are assumed to be supported within the context of each paper. For the {\name}-Med, we source data from the PubMed Central corpus via MedICaT, filtering for instances where authors explicitly state conclusions using claim-indicating verbs (e.g., \textit{show, demonstrate}, see Appendix \ref{sec:claim_indicating}). This ensures that retained examples correspond to genuine author claims grounded in the associated figure, rather than non-checkworthy descriptions. For our general science domain dataset, the {\name}-Gen, we utilize the arXiv-based domains in SciMMIR collection. 

\noindent\textbf{Claim Extraction.} Given the filtered instances obtained from the data collection stage, our goal is to extract main supported scientific claims explicitly concluded by authors. We leverage state-of-the-art models as our backbone: \texttt{google/medgemma-27b-it}~\citep{Sellergren2025MedGemma} for the medical domain and \texttt{google/gemma-3-27b-it}~\citep{GemmaTeam2024, GemmaTeam2025} for general science domain. These models were selected for their superior domain-specific reasoning capabilities, especially in medical domain. 
We provided the claim extraction model with the collected figure, caption, and specific citing paragraph, to extract supported claims corresponding to each figure-caption pair. Following prior work on text-only scientific claim verification \citep{Wadden2020a}, we apply a filtering strategy to prevent data leakage and dataset artifacts. Because captions in medical domains are often highly descriptive, we explicitly test whether the verification label can be predicted from the text alone. We discard any instance where the model successfully verifies the claim using only the caption, thereby enforcing a strict dependency on the visual evidence.\\

\vspace{-4mm}
\noindent\textbf{Perturbation Module.} To support diagnostic evaluation, we generate refuted claims by perturbing supported instances based on evidence-grounded taxonomies. While we apply standard negation protocols for the {\name}-Gen dataset following \cite{Wadden2020a, Sciver2025acl}, we utilize domain-specific perturbations to generate negatives for the medical domain. Specifically, for {\name}-Med, we define a taxonomy of seven expert-curated perturbation types, ranging from simple \textit{Status Swaps} to complex \textit{Anatomical Location Shifts}. To ensure the benchmark remains challenging for domain experts, we implement a hardness selection strategy: straightforward perturbations are randomly dropped in favor of more subtle, reasoning-intensive corruptions. This results in a dataset where refuted claims are semantically plausible but factually incorrect based on the visual evidence. Table~\ref{tab:perturbations} details the medical perturbation taxonomy and descriptions used for {\name}-Med.

\noindent\textbf{Explanation Generation and Quality Control.} To provide interpretability at scale, we formulate explanation generation as a constrained rationalization task using our backbone models. By conditioning the model on the claim and the multimodal evidence, we enforce the generation of rationales that explicitly map visual and textual features to the final verification decision. Subsequently, to guarantee the reliability of these generated instances, we deploy an LLM-as-a-Judge framework to systematically evaluate the overall claim-evidence alignment. The Judge assigns a grounding score that measures how well the scientific claim is supported by the evidence set. Based on these scores, we apply a strict rejection threshold to systematically discard weakly grounded or hallucinated instances. Experimental and reproducibility details, including prompts used for perturbation strategy, explanation rationalize, and judge settings are provided in Appendix \ref{sec:appendix_prompts_generation}, \ref{sec:llm_judge_score}.

\begin{table*}[t]
\centering
\renewcommand{\arraystretch}{1.1}
\setlength{\tabcolsep}{4.5pt} 
\small

\begin{tabular*}{\textwidth}{@{\extracolsep{\fill}} lll @{}}
\toprule
\textbf{Type} & \textbf{Description} & \textbf{Example} \\
\midrule

1. Status Swap &
Flip normal and abnormal states &
\textcolor{ForestGreen}{No evidence of X} $\rightarrow$ \textcolor{red}{Evidence of X} \\ 

2. Numeric Change &
Alter numeric values plausibly &
Tumor size \textcolor{ForestGreen}{2 cm} $\rightarrow$ \textcolor{red}{4 cm} \\

3. Attribute Flip &
Invert descriptive attributes &
\textcolor{ForestGreen}{significant edema} $\leftrightarrow$ \textcolor{red}{minimal edema} \\

4. Directional Flip &
Reverse comparative relationships &
\textcolor{ForestGreen}{$A>B$} $\leftrightarrow$ \textcolor{red}{$B>A$} \\

5. Diagnosis Swap &
Substitute primary diagnosis &
\textcolor{ForestGreen}{cancer} $\leftrightarrow$ \textcolor{red}{infection} \\

6. Location Shift &
Shift to adjacent anatomy &
\textcolor{ForestGreen}{thalamus} $\leftrightarrow$ \textcolor{red}{basal ganglia} \\

7. Certainty Shift &
Modify diagnostic confidence &
\textcolor{ForestGreen}{suspicious for} $\rightarrow$ \textcolor{red}{diagnostic of} \\
\bottomrule
\end{tabular*}
\caption{Medical perturbation taxonomy with descriptions and examples.}
\vspace{-4mm}
\label{tab:perturbations}
\end{table*}

\vspace{-3mm}
\subsection{Multi-Stage Expert Human Evaluation}
\vspace{-2mm}


\paragraph{Phase 1: Blind Inter-Annotator Agreement} To evaluate dataset reliability \citep{wu-etal-2024-scimmir, Sciver2025acl}, we conducted an evaluation with 92 verified domain experts. Following standard protocols \citep{Sciver2025acl}, we employed a blind Inter-Annotator Agreement (IAA) evaluation with stratified sampling to select 200 instances from the {\name} benchmark across both subsets, with two independent domain experts evaluating each instance. Crucially, evaluators received solely the multimodal evidence (figure and caption) and the claim, strictly without access to ground-truth labels or explanations (see Figure \ref{fig:instruction}). This blind setting resulted in an IAA of 79.6\% for the general science subset and 65.4\% for the medical subset. These results show substantial agreement for the general science and medical domains. The lower medical agreement rate directly reflects the inherent complexity and subjective nuances of interpreting specialized medical visuals than general science. Ultimately, these scores establish a realistic human baseline, demonstrating that while the claims are verifiable, the medical domain presents a remarkably challenging reasoning task.

\vspace{-3mm}
\paragraph{Phase 2: Large-Scale Alignment Validation.} In the second phase, we conducted another human intelligence task to evaluate the alignment of instances at the claim and explanation levels at scale, following the human evaluation settings of \citet{wu-etal-2024-scimmir}. Annotators evaluated a balanced subset of 1,295 claims alongside ground-truth labels. We observed a human-data agreement rate of 79.2\% for {\name}-Gen. While the general science agreement remained consistent with its blind IAA, the medical agreement increased to 84.9\% for {\name}-Med. This shows the higher cognitive load of blind label prediction compared to the alignment task. Although blind diagnosis is highly subjective, experts showed higher agreement when medical labels were provided and the human intelligence task was less cognitively demanding. Separately, experts evaluated 1,096 generated explanations to determine if the reasoning adequately justified the assigned labels. They validated the high quality and faithfulness of the explanations in 84.8\% of {\name}-Med cases and 72.6\% of {\name}-Gen cases. Comprehensive details regarding both phases of human evaluation, including annotator demographics, provided instructions, compensation rates, and the screening process\footnote{This study was approved by the Institutional Review Board (IRB).}, are available in Appendix \ref{sec:human_eval}.
\vspace{-3mm}

\paragraph{Phase 3: Statistical Reliability of Experiments.} The final phase of our quality evaluation was to confirm that our dataset serves as a reliable proxy for human reasoning. To achieve this, we established a gold standard set consisting exclusively of the human-verified instances from the second phase. We then conducted a statistical audit by comparing the baseline performance scores and rankings of models on this gold subset against their scores and rankings on the full test set. We observed highly significant correlations between the gold set and the full test set across both domains. Specifically, {\name}-Gen achieved a Pearson $r = 0.956$ and a Spearman $\rho = 0.943$ ($p = 0.005$), while {\name}-Med achieved a Pearson $r = 0.917$ and a Spearman $\rho = 0.899$ ($p = 0.015$). This statistical correlation proves the large scale test set reliably reflects expert judgment. Furthermore, it ensures that potential residual bias or pipeline artifacts do not skew evaluation results thus guaranteeing overall benchmark reliability.

\noindent\textbf{Dataset Statistics.} {\name} contains 469,264 claims from PubMed and 16 arXiv domains referenced in Appendix \ref{sec:arxiv_categories}. The benchmark allocates 15,953 instances to {\name} Med and 453,311 instances to {\name} Gen. We enforce a strict paper level 60/20/20 split across train validation and test partitions to prevent data leakage. Additionally every claim is paired with an explanation of roughly 100 words detailing how the multimodal evidence justifies the consistency label.

\begin{table}[htbp]
\centering
\renewcommand{\arraystretch}{1}
\setlength{\tabcolsep}{3pt} 
\small

\begin{tabular}{lcccc}
\toprule
\multicolumn{1}{c}{\bf Model} &
\multicolumn{2}{c}{\bf \name-Med} &
\multicolumn{2}{c}{\bf \name-Gen} \\
\cmidrule{2-3}
\cmidrule{4-5}
\multicolumn{1}{c}{} &
\multicolumn{1}{c}{\bf Macro-F1} &
\multicolumn{1}{c}{\bf Micro-F1} &
\multicolumn{1}{c}{\bf Macro-F1} &
\multicolumn{1}{c}{\bf Micro-F1} \\
\midrule

\multicolumn{5}{c}{\bf Open-source Vision-Language Models} \\
\midrule
Phi-4-Multimodal        & 65.8 & 72.3 & 45.7 & 47.4 \\
Pixtral-12B             & 69.6 & 75.9 & 41.9 & 59.2 \\
LLaMA-3.2-11B-Vision    & 69.3 & 70.8 & 53.8 & 53.8 \\
Mistral-Small-3.1-24B   & 71.4 & 72.7 & 63.1 & 64.1 \\
InternVL3-8B            & 69.6 & \underline{76.8} & \underline{63.8} & \underline{70.3} \\
Qwen2.5-VL-7B           & 70.4 & 71.3 & \textbf{68.1} & \textbf{70.4}\\

\midrule
\multicolumn{5}{c}{\bf Closed-source Models} \\
\midrule
GPT-5-mini              & 67.2 & 67.3 & 59.9 & 59.3 \\
GPT-4o-mini             & \textbf{73.0} & \textbf{77.5} & 56.6 & 65.0 \\

\midrule
\multicolumn{5}{c}{\bf Task-specific Claim-Verification Models} \\
\midrule
LVLM4FV                 & 38.6 & 58.8 & 49.9 & 63.5 \\
MOCHEG                  & \underline{71.8} & 76.3 & 52.1 & 60.5 \\

\midrule
\multicolumn{5}{c}{\bf Domain-Specific Medical Models} \\
\midrule
LLaVA-Med               & 51.8 & 64.3 & --    & --    \\
HuatuoGPT-Vision-7B     & 60.0 & 72.0 & --    & --    \\

\bottomrule
\end{tabular}
\caption{Comparison of model performance (\%) on the {\name} test sets. Dashes indicate models not evaluated on a subset. Best scores in \textbf{bold}, second best \underline{underlined}.}
\label{tab:scifc_merged_label}
\end{table}
\vspace{-5mm}

\begin{table*}[t]
\centering
\renewcommand{\arraystretch}{1.1}

\resizebox{\textwidth}{!}{%
\begin{tabular}{lcccccccc}
\toprule
\textbf{Model} &
\multicolumn{2}{c}{\textbf{Low-Complexity}} &
\multicolumn{3}{c}{\textbf{Intermediate-Complexity}} &
\multicolumn{2}{c}{\textbf{High-Complexity}} &
\textbf{Avg.} \\
\cmidrule(lr){2-3}
\cmidrule(lr){4-6}
\cmidrule(lr){7-8}
 &
\textbf{Certainty} &
\textbf{Status} &
\textbf{Diagnosis} &
\textbf{Attribute} &
\textbf{Numeric} &
\textbf{Location} &
\textbf{Directional} &
\\
\midrule

InternVL3-8B &
50.0 & 60.0 & 37.4 & 35.3 & 37.7 & 20.5 & 21.4 &
37.5 $\pm$ 13.2 \\

Phi-4-Multimodal &
66.7 & 63.4 & 32.7 & 34.7 & 44.9 & 19.4 & 26.8 &
41.2 $\pm$ 16.7 \\

Pixtral-12B &
66.7 & 63.1 & 40.5 & 38.9 & 38.9 & 26.0 & 19.6 &
42.0 $\pm$ 16.2 \\

LLaMA-3.2-11B-Vision &
83.3 & 86.7 & 60.7 & 65.8 & 68.9 & 42.5 & 53.6 &
65.9 $\pm$ 14.5 \\

Mistral-Small-3.1-24B &
83.3 & 84.9 & 65.0 & 70.5 & 78.4 & 63.0 & 50.0 &
70.7 $\pm$ 11.6 \\

Qwen2.5-VL-7B &
83.3 & 85.8 & 74.8 & 74.8 & 79.0 & 61.6 & 48.2 &
72.5 $\pm$ 12.3 \\

\bottomrule
\end{tabular}
}

\caption{Micro-F1 (\%) across medical perturbations on the {\name}-Med test set.}
\vspace{-4mm}
\label{tab:medicat_perturbations}
\end{table*}

\begin{table*}[t]
\centering
\renewcommand{\arraystretch}{1}
\setlength{\tabcolsep}{4.5pt} 
\small

\begin{tabular}{lcccccc}
\toprule
\textbf{Model} 
& \multicolumn{3}{c}{\textbf{\name-Med}} 
& \multicolumn{3}{c}{\textbf{\name-Gen}} \\
\cmidrule(lr){2-4}
\cmidrule(lr){5-7}
 & \textbf{ROUGE-L} & \textbf{BLEU-2} & \textbf{METEOR}
 & \textbf{ROUGE-L} & \textbf{BLEU-2} & \textbf{METEOR} \\
\midrule

Phi-4-MM-Instruct     & 31.05 & 30.60 & 35.70 &  27.49 & 24.56 & 32.10 \\
Qwen2.5-VL-7B         & 31.76 & \underline{32.48} & \underline{39.11} & 27.48 & 21.24  & 27.72 \\
Pixtral-12B           & \textbf{32.95} & 31.14 & 35.96 & \textbf{32.12} & \textbf{28.44} & \underline{33.93} \\
InternVL3-8B          & 31.64 & \textbf{32.62} & \textbf{39.39} & \underline{29.60} & 24.66 & \textbf{36.92} \\
\midrule

GPT-4o-mini           & \underline{32.60} & 30.81 & 36.11 & 27.92 & 24.68 & 30.23\\
GPT-5-mini            & 30.24 & 26.61 & 32.94 & 27.98 & \underline{24.83} & 30.41 \\
\midrule
Llava-Med             & 25.85 & 15.89 & 24.56 & --    & --    & --    \\
HuatuoGPT-Vision-7B   & 30.17 & 20.80 & 31.52 & --    & --    & --    \\

\bottomrule
\end{tabular}

\caption{Explanation generation (\%) on {\name}-Med and {\name}-Gen test sets.}
\label{tab:expl_med_vs_gen}
\vspace{-4mm}
\end{table*}

\begin{table}[htbp]
\centering
\renewcommand{\arraystretch}{1.1}
\setlength{\tabcolsep}{2pt} 
\small

\begin{tabular}{clccccc}
\toprule
&\textbf{Model} 
& \textbf{Entailment} 
& \textbf{Relevance} 
& \textbf{Correctness} 
& \textbf{Completeness} 
& \textbf{Clarity} \\
\midrule
\multirow{6}{*}{\rotatebox{90}{{\name}-Med}} &
Qwen2.5-VL-7B-Instruct      & 7.76 & 9.50 & 7.19 & 7.48 & 8.73 \\
& InternVL3-8B               & 8.19 & 9.53 & 7.70 & 7.66 & 8.87 \\
& Phi-4-Multimodal-Instruct  & 8.10 & 9.33 & 7.39 & 7.30 & 8.65 \\
& Pixtral-12B                & 8.51 & 9.38 & 7.83 & 7.32 & 8.86 \\\cdashline{2-7}
& GPT-4o-mini                & \underline{8.87} & \underline{9.67} & \underline{8.46} & \underline{7.82} & \underline{9.14} \\
& GPT-5-mini                 & \textbf{9.47} & \textbf{9.94} & \textbf{9.43} & \textbf{8.86} & \textbf{9.41} \\

\midrule
\multirow{6}{*}{\rotatebox{90}{{\name}-Gen}} &
Qwen2-VL-7B-Instruct       & 4.81 & 6.56 & 3.99 & 4.39 & 6.80 \\
& InternVL3-8B               & 6.53 & 7.98 & 5.72 & 6.02 & 7.86 \\
& Phi-4-Multimodal-Instruct  & 5.15 & 7.18 & 4.79 & 5.02 & 7.73 \\
& Pixtral-12B                & 6.42 & 8.30 & 5.58 & 5.66 & 8.08 \\\cdashline{2-7}
& GPT-4o-mini                & \textbf{7.39} & \textbf{9.11} & \textbf{7.01} & \textbf{6.89} & \textbf{8.84} \\
& GPT-5-mini                 & \underline{7.22} & \underline{8.93} & \underline{6.90} & \underline{6.81} & \underline{8.83} \\

\bottomrule
\end{tabular}
\caption{LLM-as-a-Judge explanation quality on {\name}. Scores average entailment, relevance, correctness, completeness, and clarity on a 1-to-10 scale ($\uparrow$).}
\vspace{-6mm}
\label{tab:exp_llm_as_a_judge}
\end{table}
\section{Experiment}

\vspace{-3mm}
\subsection{Experimental Setup}
\label{sec:experiment_setup}
\vspace{-2mm}

\noindent\textbf{Baselines.} We evaluate 12 models across four categories. The baselines include closed source foundation models GPT-5 mini \citep{openai2025gpt5} and GPT-4o mini \citep{openai2024gpt4o} alongside open source vision language models \citep{Sciver2025acl} such as Qwen2.5 VL 7B \citep{qwen25vl2024} Pixtral 12B \citep{pixtral12b2024} Mistral Small 3.1 24B \citep{mistral2025small31} Phi 4 Mini \citep{phi4mini2025} InternVL3 8B \citep{chen2024expanding, wang2024mpo, chen2024far, chen2024internvl} and LLaMA 3.2 Vision \citep{meta2024llama3}. We test general domain claim verification methods MOCHEG \citep{MOCHEG} and LVLM4FV \citep{cikm_Tahmasebi24} across both datasets while restricting medical models LLaVA Med \citep{Li2023LLaVAmed} and HuatuoGPT Vision 7B \citep{chen2024huatuogptvisioninjectingmedicalvisual} exclusively to {\name}-Med. To enforce strictly grounded evaluation we adapted the general domain methods by fine tuning MOCHEG directly on {\name} to bypass retrieval and restricting LVLM4FV to the single gold image caption pair. Baseline prompts are detailed in Appendix \ref{appendix:baseline_prompts}.

\noindent\textbf{Evaluation Metrics.} We evaluate baselines on claim verification and explanation quality. Following \citep{ZhangLee2024} we report Macro F1 and Micro F1 scores for verification. For explanations we calculate N-gram metrics (BLEU-2, ROUGE-L, and METEOR) alongside an LLM-as-a-Judge measuring correctness, relevance, completeness, entailment, and clarity. Finally, evaluation by domain experts assesses explanation faithfulness and identifies hallucination cases.
\vspace{-2mm}

\begin{table*}[t]
\centering
\renewcommand{\arraystretch}{1.1}

\resizebox{\textwidth}{!}{%
\begin{tabular}{lccccccccccc}
\toprule
\textbf{Model} &
\textbf{cs} &
\textbf{econ} &
\textbf{eess} &
\textbf{gr-qc} &
\textbf{hep} &
\textbf{math} &
\textbf{physics} &
\textbf{q-bio} &
\textbf{q-fin} &
\textbf{quant-ph} &
\textbf{stat} \\
\midrule

Phi-4-Multimodal &
45.2 & 45.0 & 45.4 & 45.4 & 46.0 & 44.8 & 46.5 & 46.5 & 43.0 & 47.1 & 44.5  \\

Pixtral-12B &
42.4 & 40.3 & 41.2 & 42.1 & 41.2 & 42.9 & 41.4 & 41.5 & 41.8 & 41.2 & 41.2 \\

Mistral-Small-3.1-24B &
51.7 & 44.3 & 53.0 & 53.4 & 53.0 & 50.8 & 54.0 & 53.7 & 46.1 & 53.6 & 52.3  \\

LLaMA-3.2-11B-Vision &
54.5 & 47.7 & 55.7 & 52.2 & 52.4 & 48.8 & 53.8 & 52.2 & 52.9 & 53.1 & 50.3 \\

InternVL3-8B &
67.2 & 65.6 & 65.5 & 61.4 & 59.3 & 61.8 & 58.1 & 63.3 & 70.3 & 60.9 & 60.8 \\

Qwen2.5-VL-7B &
70.3 & 70.4 & 70.4 & 67.7 & 65.0 & 65.4 & 65.9 & 71.0 & 72.9 & 66.7 & 65.6 \\

\bottomrule
\end{tabular}
}

\caption{
Domain-wise macro-F1 (\%) performance on the {\name}-Gen test set.
}
\label{tab:domain_performance}
\vspace{-2mm}
\end{table*}











\begin{table}[htbp]
\centering
\renewcommand{\arraystretch}{1}
\setlength{\tabcolsep}{3pt} 
\small

\begin{tabular}{llcccc}
\toprule
\multicolumn{1}{c}{\bf Model} &
\multicolumn{1}{c}{\bf Setting} &
\multicolumn{2}{c}{\bf \name-Med} &
\multicolumn{2}{c}{\bf \name-Gen} \\
\cmidrule{3-4}
\cmidrule{5-6}
\multicolumn{1}{c}{} &
\multicolumn{1}{c}{} &
\multicolumn{1}{c}{\bf Macro-F1} &
\multicolumn{1}{c}{\bf Micro-F1} &
\multicolumn{1}{c}{\bf Macro-F1} &
\multicolumn{1}{c}{\bf Micro-F1} \\
\midrule

\multirow{3}{*}{Qwen2.5-VL-7B}
& w/o image  & 68.1 & 71.0 & 52.2 & 53.2 \\
& w/o text   & 64.0 & 65.4 & 50.4 & 50.7 \\
& Multimodal & 70.4 & 71.3 & 68.1 & 70.4 \\
\midrule

\multirow{3}{*}{GPT-4o-mini}
& w/o image  & 56.4 & 56.5 & 43.6 & 46.6 \\
& w/o text   & 50.9 & 51.1 & 56.4 & 64.8 \\
& Multimodal & 73.0 & 77.5 & 56.6 & 65.0 \\
\midrule

\multirow{3}{*}{MOCHEG}
& w/o image  & 67.0 & 74.6 & 55.3 & 60.8 \\
& w/o text   & 67.9 & 69.2 & 50.6 & 58.9 \\
& Multimodal & 71.8 & 76.3 & 52.1 & 60.5 \\
\bottomrule
\end{tabular}
\caption{Ablation study of modality contributions. 
Results are in percentage.}

\label{tab:scifc_modality_ablation}
\vspace{-5mm}
\end{table}

\vspace{-2mm}
\subsection{Claim Verification and Diagnostics}
\vspace{-2mm}
\noindent\textbf{Claim Verification.} 
Table~\ref{tab:scifc_merged_label} presents the performance of the baselines on the claim verification task. GPT-4o-mini achieves the highest performance on the medical subset with a 73.0\% Macro-F1. In contrast, open-source models demonstrate stronger generalization across the broader scientific domains; specifically, Qwen2.5-VL-7B achieves a 68.1\% Macro-F1 on the general science subset. Overall, the results highlight a distinct performance disparity across domains: the leading closed-source model excels in specialized medical verification, while top open-source models exhibit greater robustness on general science claims.

\noindent\textbf{Reasoning Complexity.}
Utilizing {\name}-Med, we evaluate open-source vision-language models on our medical perturbation taxonomy to identify failure modes in Table \ref{tab:medicat_perturbations}. We group perturbations into three categories consisting of low, intermediate, and high complexity. The best performing open-source model, Qwen2.5-VL-7B, shows performance decrease from 85.8 on the low-complexity normal-to-abnormal status swapping to 61.6 on the high-complexity anatomical location shift perturbation. All open-source baselines show distinct performance across these complexity levels, highlighting that current models specifically struggle with highly complex reasoning tasks in medical settings. 

\noindent\textbf{Domain Specificity.} Using {\name}-Gen, we conducted extensive experiments across arXiv sub-domains to evaluate model performance (Table~\ref{tab:domain_performance}). Performance is noticeably non-uniform across scientific fields. The best-performing model, Qwen2.5-VL-7B, performs exceptionally well in Quantitative Finance (72.9\%) and Computer Science (70.3\%). In contrast, models struggle significantly in more abstract domains. For example, InternVL3-8B sees its absolute lowest performance in general Physics (58.1\%) and High-Energy Physics (59.3\%). Overall, baselines consistently underperform in abstract fields such as Mathematics and Physics, where diagrammatic reasoning differs fundamentally from standard plots.
\vspace{-4mm}
\subsection{Task Learnability via Fine-Tuning}
\label{sec:task_learnability}
\vspace{-2mm}
\noindent\textbf{Training Recipe.} To separate inherent model capacity from zero-shot prompt engineering techniques and establish achievable performance upper bounds, we conducted supervised fine-tuning (SFT) on three foundational vision-language models: InternVL3-8B, LLaMA-3.2-11B-Vision, and Qwen2.5-VL-7B. We designed an efficient and reproducible training recipe using LoRA \citep{hu2022lora} applied to all linear attention and MLP modules, with hyperparameters set to $r=32$, $\alpha=64$, and a dropout rate of $0.05$. The models were optimized using a learning rate of $5 \times 10^{-5}$. Furthermore, to process the high-fidelity visual evidence inherent to scientific domains (e.g., complex radiology scans) without exceeding GPU memory constraints, we capped the maximum image resolution at 401,408 total pixels during training for the {\name}-Med subset.

\begin{figure*}[t]
    \centering
    \begin{tabular}{ccc}
        \includegraphics[width=0.31\textwidth]{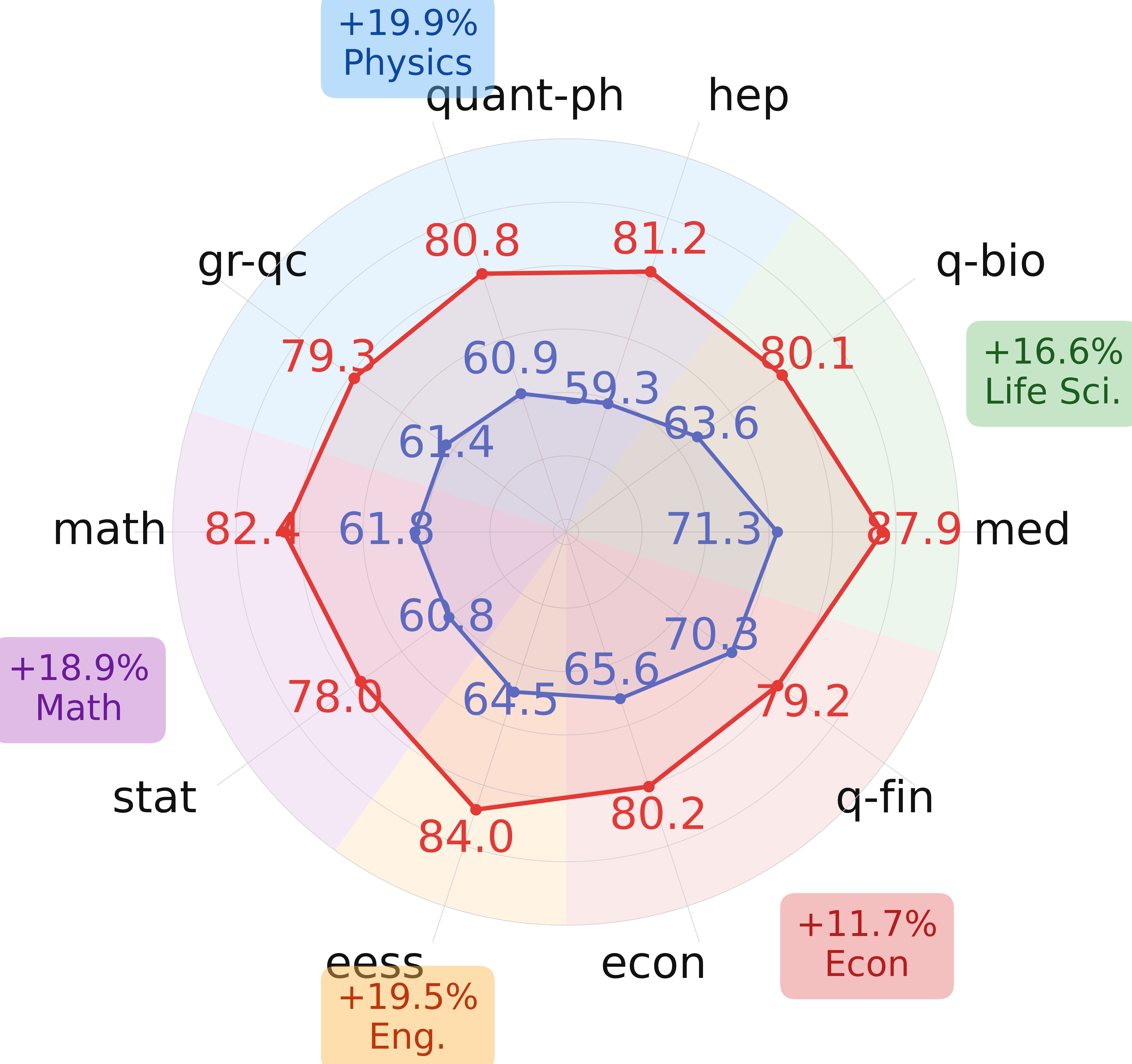} & 
        \includegraphics[width=0.31\textwidth]{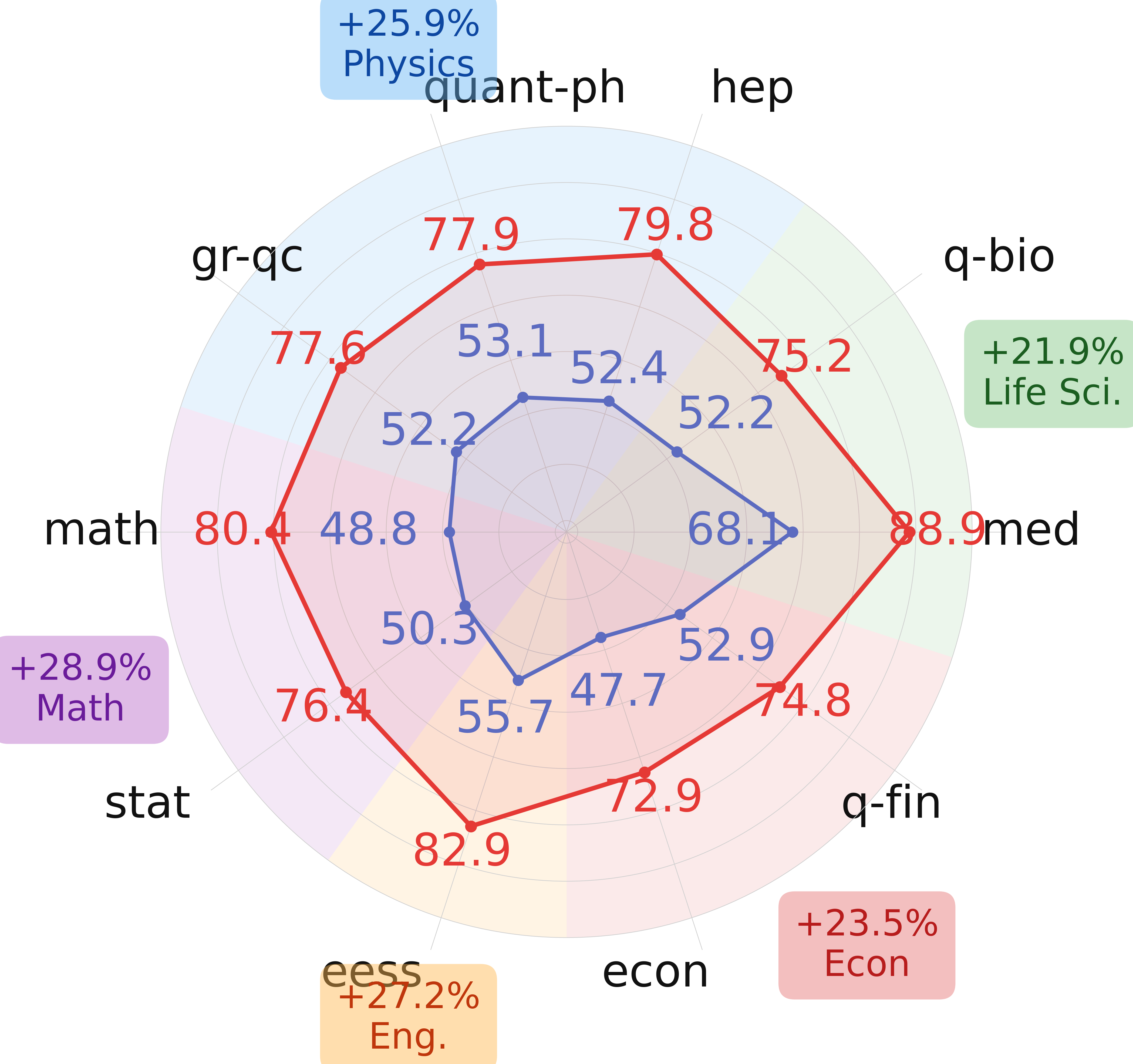} &
        \includegraphics[width=0.31\textwidth]{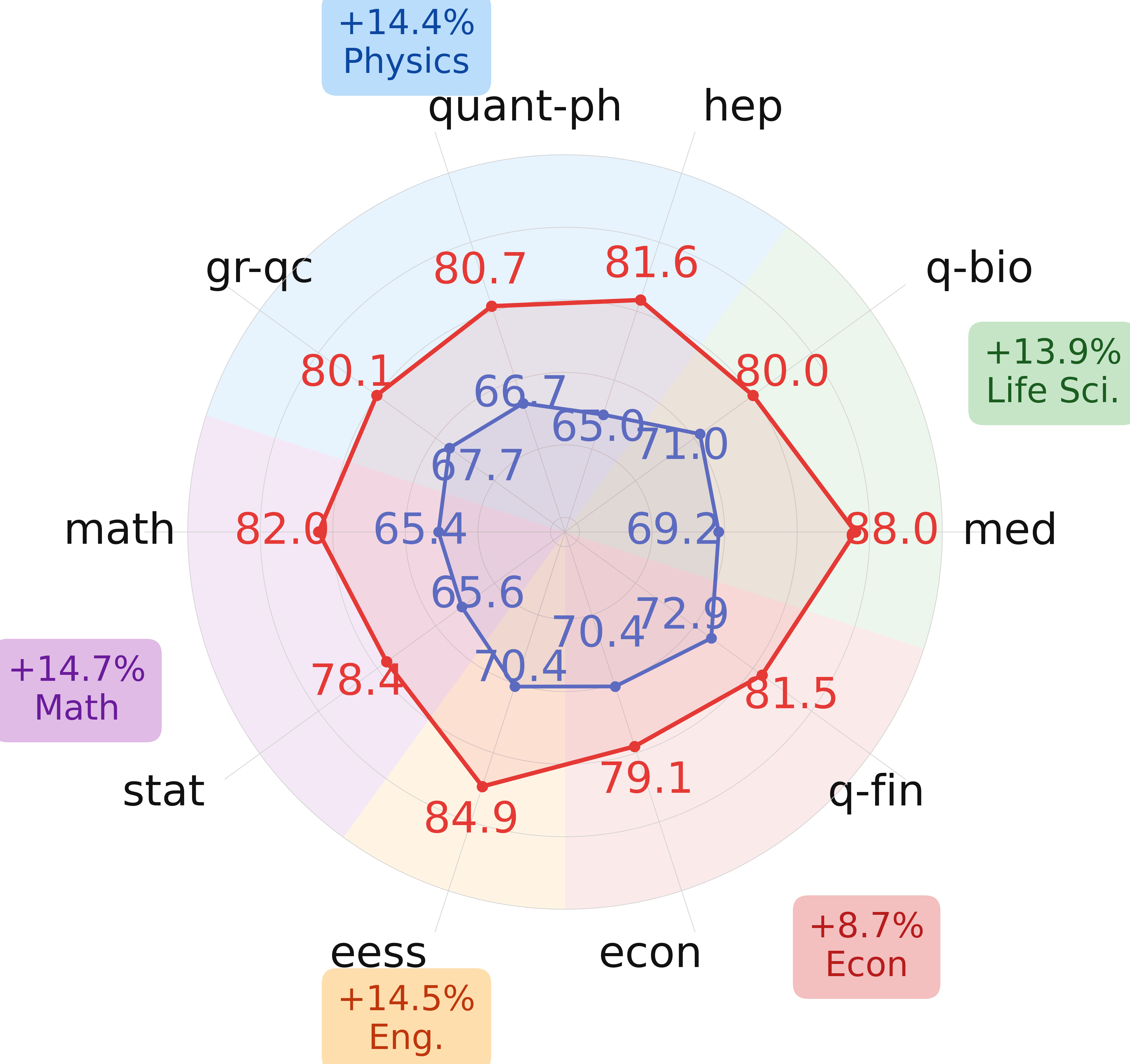} \\
        \small (a) InternVL3-8B & \small (b) Llama-3.2-11B & \small (c) Qwen2.5-VL-7B \\
    \end{tabular}
\caption{Radar plots comparing zero-shot baselines (blue) and SFT claim verification performance (red) across diverse scientific domains on the {\name} benchmark.}
    \label{fig:radar_results_1x3}
    \vspace{-2mm}
\end{figure*}

\noindent\textbf{Task Learnability and Performance Gains.} The fine-tuning results reveal a consistent learning signal across all evaluated models and scientific disciplines, as shown in Figure~\ref{fig:radar_results_1x3}. While zero-shot baselines struggled with abstract diagrammatic reasoning, post-SFT performance shows high improvements across the board. Most notably, LLaMA-3.2-11B-Vision (Figure~\ref{fig:radar_results_1x3}b) achieved high gains in difficult domains, up to +28.9\% in Mathematics, +27.2\% in Engineering (EESS), and +25.9\% in Physics. Similarly, InternVL3-8B and Qwen2.5-VL-7B highlights broad performance lifts of up to +19.9\% and +14.7\% in abstract fields, alongside robust improvements ranging from +13.9\% to +21.9\% in the Life Sciences (q-bio) and Medical subsets. These substantial gains confirm that {\name} provides the domain-specific cross-modal alignment necessary to teach foundation models complex scientific verification.

\vspace{-3mm}

\subsection{Explanation Evaluation} 
\label{sec:explanation_evaluation}

\vspace{-2mm}
\noindent\textbf{Surface-Level Fluency and Semantic Validity.} We evaluated explanation quality using N-gram metrics presented in Table~\ref{tab:expl_med_vs_gen} to measure lexical overlap alongside an LLM-as-a-Judge framework via DeepSeek-3.2 detailed in Table~\ref{tab:exp_llm_as_a_judge} to measure semantic reasoning. Open-source models show strong surface-level fluency by frequently matching closed-source models. For example, Pixtral-12B achieves the highest ROUGE-L scores across both domains with 32.95\% and 32.12\%, respectively. However, scientific explanations require logical validity beyond lexical similarity. Our semantic evaluation shows a fundamental shift where closed-source models significantly outperform open-source models in reasoning. For instance, on the {\name}-Med subset, GPT-5-mini achieves the highest scores across all dimensions, including 9.43 in Correctness and 8.86 in Completeness, compared to InternVL3-8B's 7.70 and 7.66. This highlights that while open-source models can generate fluent text while they struggle with incomplete reasoning logic when justifying claim. We provide further detail regarding the prompt supplied to the language model judge in Appendix~\ref{sec:appendix_llm_judge}.

\noindent\textbf{Expert Evaluation.} To empirically validate this divergence between surface level fluency and semantic validity we conducted a blind human evaluation. We recruited 20 domain experts via Prolific to blindly evaluate 240 explanations sampled from GPT-5-mini, Pixtral-12B, and Qwen2.5-VL-7B. Medical annotators identified numerical scaling errors including hallucinated biological ranges in open source outputs while general science annotators found linguistic artifacts (see case study Appendix~\ref{sec:appendix_case_study_artifacts}). Furthermore experts rated overall explanation faithfulness on a scale of 1 to 5 where GPT-5-mini scored the highest with 3.54 followed by Pixtral-12B with 3.41 and Qwen2.5-VL-7B with 3.38. These expert audits highlight that although open source models can generate lexically fluent explanations their reasoning often lacks the critical logical faithfulness necessary for justification.

\vspace{-4mm}
\subsection{Modality Ablation} 
\vspace{-3mm}
To evaluate the necessity of multimodal reasoning we conducted an ablation study detailed in Table~\ref{tab:scifc_modality_ablation}. The full multimodal setting consistently achieves the highest Macro-F1 scores. For instance the GPT-4o-mini score on the medical subset drops from 73.0\% to 56.4\% without images and 50.9\% without text. Similarly Qwen2.5-VL-7B performance falls from 68.1\% to 52.2\% without visual inputs in the general science domain. Conversely the fine tuned MOCHEG model presents an anomaly on the general science subset where its text only baseline outperforms the multimodal setting with 55.3\% compared to 52.1\%. We hypothesize that traditional fusion architectures struggle to integrate diverse scientific imagery resulting in representation interference that degrades their text processing capabilities. Modern vision language models however effectively synthesize both streams proving that advanced joint reasoning is essential for this benchmark.

\vspace{-4mm}
\subsection{Case Study} 
\vspace{-3mm}
As illustrated in Figure~\ref{fig:data_samples} our qualitative analysis reveals two critical failure modes in top models. First vision language models struggle with fine grained spatial reasoning. When challenged with an anatomical positioning shift InternVL3-8B fails to distinguish between the stomach and the duodenum in an X-ray and incorrectly supports a refuted claim. Second these models frequently exhibit visual bypassing where they ignore visual evidence in favor of parametric memory. For instance when evaluating an ASR architecture diagram a model hallucinates a six layer structure despite the figure clearly indicating twelve transformer blocks. These examples show that models often bypass rigorous visual verification when faced with subtle anatomical shifts or complex scientific charts.


\vspace{-4mm}
\section{Conclusion}
\vspace{-3mm}
We introduce {\name}, a large-scale multimodal and multidomain dataset for scientific claim consistency evaluation, featuring 469K visually grounded claims with extensive expert validation. Our evaluations show that robust verification and explanation across complex scientific and medical domains remain challenging for state-of-the-art VLMs. By highlighting these limitations, {\name} supports future advances in cross-modal scientific reasoning and the development of trustworthy multimodal AI.

\bibliography{colm2026_conference}

@inproceedings{Wadden2020a,
    author    = {Wadden, David and Lin, Shanchuan and Lo, Kyle and Wang, Lucy Lu and van Zuylen, Madeleine and Cohan, Arman and Hajishirzi, Hannaneh},
    title     = {Fact or fiction: Verifying scientific claims},
    editor    = {Webber, Bonnie and Cohn, Trevor and He, Yulan and Liu, Yang},
    booktitle = {Proceedings of the 2020 Conference on Empirical Methods in Natural Language Processing (EMNLP)},
    pages     = {7534--7550},
    publisher = {Association for Computational Linguistics},
    year      = {2020},
    doi       = {10.18653/v1/2020.emnlp-main.609}
}

@inproceedings{ZhangLee2024,
    title = "{CORRECT}: Context- and Reference-Augmented Reasoning and Prompting for Fact-Checking",
    author = "Zhang, Delvin Ce  and
      Lee, Dongwon",
    editor = "Chiruzzo, Luis  and
      Ritter, Alan  and
      Wang, Lu",
    booktitle = "Proceedings of the 2025 Conference of the Nations of the Americas Chapter of the Association for Computational Linguistics: Human Language Technologies (Volume 1: Long Papers)",
    month = apr,
    year = "2025",
    address = "Albuquerque, New Mexico",
    publisher = "Association for Computational Linguistics",
    url = "https://aclanthology.org/2025.naacl-long.154/",
    doi = "10.18653/v1/2025.naacl-long.154",
    pages = "3007--3019",
    ISBN = "979-8-89176-189-6"
}

@inproceedings{Aly2021,
    title = "The Fact Extraction and {VER}ification Over Unstructured and Structured information ({FEVEROUS}) Shared Task",
    author = "Aly, Rami  and
      Guo, Zhijiang  and
      Schlichtkrull, Michael Sejr  and
      Thorne, James  and
      Vlachos, Andreas  and
      Christodoulopoulos, Christos  and
      Cocarascu, Oana  and
      Mittal, Arpit",
    editor = "Aly, Rami  and
      Christodoulopoulos, Christos  and
      Cocarascu, Oana  and
      Guo, Zhijiang  and
      Mittal, Arpit  and
      Schlichtkrull, Michael  and
      Thorne, James  and
      Vlachos, Andreas",
    booktitle = "Proceedings of the Fourth Workshop on Fact Extraction and VERification (FEVER)",
    month = nov,
    year = "2021",
    address = "Dominican Republic",
    publisher = "Association for Computational Linguistics",
    url = "https://aclanthology.org/2021.fever-1.1/",
    doi = "10.18653/v1/2021.fever-1.1",
    pages = "1--13"
}

@inproceedings{Lal2025MuSciClaims,
    title = "{M}u{S}ci{C}laims: Multimodal Scientific Claim Verification",
    author = "Lal, Yash Kumar  and
      Bandham, Manikanta  and
      Hasan, Mohammad Saqib  and
      Kashi, Apoorva  and
      Koupaee, Mahnaz  and
      Balasubramanian, Niranjan",
    editor = "Inui, Kentaro  and
      Sakti, Sakriani  and
      Wang, Haofen  and
      Wong, Derek F.  and
      Bhattacharyya, Pushpak  and
      Banerjee, Biplab  and
      Ekbal, Asif  and
      Chakraborty, Tanmoy  and
      Singh, Dhirendra Pratap",
    booktitle = "Proceedings of the 14th International Joint Conference on Natural Language Processing and the 4th Conference of the Asia-Pacific Chapter of the Association for Computational Linguistics",
    month = dec,
    year = "2025",
    address = "Mumbai, India",
    publisher = "The Asian Federation of Natural Language Processing and The Association for Computational Linguistics",
    url = "https://aclanthology.org/2025.ijcnlp-long.175/",
    pages = "3285--3307",
    ISBN = "979-8-89176-298-5"
}

@techreport{GemmaTeam2025,
    author    = {{Gemma Team}},
    title     = {Gemma 3 technical report},
    institution = {Google DeepMind},
    year      = {2025}
}

@inproceedings{Wang2023Covid,
    author    = {Wang, Gengyu and Harwood, Kate and Chillrud, Lawrence and Ananthram, Amith and Subbiah, Melanie and Mckeown, Kathleen},
    title     = {Check-COVID: Fact-checking COVID-19 news claims with scientific evidence},
    booktitle = {Findings of the Association for Computational Linguistics: ACL 2023},
    pages     = {14114--14127},
    year      = {2023}
}

@inproceedings{Akhtar2024ChartCheck,
    author    = {Akhtar, Mubashara and Subedi, Nikesh and Gupta, Vivek and Tahmasebi, Sahar and Cocarascu, Oana and Simperl, Elena},
    title     = {ChartCheck: Explainable fact-checking over real-world chart images},
    booktitle = {Findings of the Association for Computational Linguistics: ACL 2024},
    pages     = {13921--13937},
    address   = {Bangkok, Thailand},
    publisher = {Association for Computational Linguistics},
    year      = {2024}
}

@inproceedings{Sciver2025acl,
    author    = {Wang, Chengye and Shen, Yifei and Kuang, Zexi and Cohan, Arman and Zhao, Yilun},
    title     = {SciVer: Evaluating Foundation Models for Multimodal Scientific Claim Verification},
    booktitle = {Proceedings of the 63rd Annual Meeting of the Association for Computational Linguistics, ACL 2025},
    pages     = {8562–8579},
    year      = {2025}
}

@inproceedings{Li2023LLaVAmed,
  author    = {Li, Chunyuan and Wong, Cliff and Zhang, Sheng and Usuyama, Naoto and Liu, Haotian and Yang, Jianwei and Naumann, Tristan and Poon, Hoifung and Gao, Jianfeng},
  title     = {LLaVA-Med: Training a Large Language-and-Vision Assistant for Biomedicine in One Day},
  booktitle = {Proceedings of the 37th International Conference on Neural Information Processing Systems (NeurIPS)},
  year      = {2023},
  pages     = {28541--28564},
  articleno = {1240},
}

@article{Sellergren2025MedGemma,
  title     = {MedGemma Technical Report},
  author    = {Sellergren, Andrew and Kazemzadeh, Sahar and Jaroensri, Tiam and Kiraly, Atilla and Traverse, Madeleine and Kohlberger, Timo and Xu, Shawn and Jamil, Fayaz and Hughes, C{\'\i}an and Lau, Charles and Chen, Justin and Mahvar, Fereshteh and Yatziv, Liron and Chen, Tiffany and Sterling, Bram and Baby, Stefanie Anna and Baby, Susanna Maria and Lai, Jeremy and Schmidgall, Samuel and Yang, Lu and Chen, Kejia and Bjornsson, Per and Reddy, Shashir and Brush, Ryan and Philbrick, Kenneth and Asiedu, Mercy and Mezerreg, Ines and Hu, Howard and Yang, Howard and Tiwari, Richa and Jansen, Sunny and Singh, Preeti and Liu, Yun and Azizi, Shekoofeh and Kamath, Aishwarya and Ferret, Johan and Pathak, Shreya and Vieillard, Nino and Merhej, Ramona and Perrin, Sarah and Matejovicova, Tatiana and Ram{\'e}, Alexandre and Riviere, Morgane and Rouillard, Louis and Mesnard, Thomas and Cideron, Geoffrey and Grill, Jean-Bastien and Ramos, Sabela and Yvinec, Edouard and Casbon, Michelle and Buchatskaya, Elena and Alayrac, Jean-Baptiste and Lepikhin, Dmitry and Feinberg, Vlad and Borgeaud, Sebastian and Andreev, Alek and Hardin, Cassidy and Dadashi, Robert and Hussenot, L\'{e}onard and Joulin, Armand and Bachem, Olivier and Matias, Yossi and Chou, Katherine and Hassidim, Avinatan and Goel, Kavi and Farabet, Clement and Barral, Joelle and Warkentin, Tris and Shlens, Jonathon and Fleet, David and Cotruta, Victor and Sanseviero, Omar and Martins, Gus and Kirk, Phoebe and Rao, Anand and Shetty, Shravya and Steiner, David F. and Kirmizibayrak, Can and Pilgrim, Rory and Golden, Daniel and Yang, Lin},
  journal   = {arXiv preprint},
  volume    = {arXiv:2507.05201},
  year      = {2025},
  url       = {https://arxiv.org/abs/2507.05201}
}

@techreport{GemmaTeam2024,
  title        = {Gemma: Open Models Based on Gemini Research and Technology},
  author       = {Mesnard, Thomas and Hardin, Cassidy and Dadashi, Robert and Bhupatiraju, Surya and Pathak, Shreya and Sifre, Laurent and Rivi{\`e}re, Morgane and Kale, Mihir Sanjay and Love, Juliette and Tafti, Pouya and Hussenot, L{\'e}onard and Sessa, Pier Giuseppe and Chowdhery, Aakanksha and Roberts, Adam and Barua, Aditya and Botev, Alex and Castro-Ros, Alex and Slone, Ambrose and H{\'e}liou, Am{\'e}lie and Tacchetti, Andrea and Bulanova, Anna and Paterson, Antonia and Tsai, Beth and Shahriari, Bobak and Le Lan, Charline and Choquette-Choo, Christopher A. and Crepy, Cl{\'e}ment and Cer, Daniel and Ippolito, Daphne and Reid, David and Buchatskaya, Elena and Ni, Eric and Noland, Eric and Yan, Geng and Tucker, George and Muraru, George-Christian and Rozhdestvenskiy, Grigory and Michalewski, Henryk and Tenney, Ian and Grishchenko, Ivan and Austin, Jacob and Keeling, James and Labanowski, Jane and Lespiau, Jean-Baptiste and Stanway, Jeff and Brennan, Jenny and Chen, Jeremy and Ferret, Johan and Chiu, Justin and Mao-Jones, Justin and Lee, Katherine and Yu, Kathy and Millican, Katie and Sjoesund, Lars Lowe and Lee, Lisa and Dixon, Lucas and Reid, Machel and Miku{\l}a, Maciej and Wirth, Mateo and Sharman, Michael and Chinaev, Nikolai and Thain, Nithum and Bachem, Olivier and Chang, Oscar and Wahltinez, Oscar and Bailey, Paige and Michel, Paul and Yotov, Petko and Chaabouni, Rahma and Comanescu, Ramona and Jana, Reena and Anil, Rohan and McIlroy, Ross and Liu, Ruibo and Mullins, Ryan and Smith, Samuel L. and Borgeaud, Sebastian and Girgin, Sertan and Douglas, Sholto and Pandya, Shree and Shakeri, Siamak and De, Soham and Klimenko, Ted and Hennigan, Tom and Feinberg, Vlad and Stokowiec, Wojciech and Chen, Yu-hui and Ahmed, Zafarali and Gong, Zhitao and Warkentin, Tris and Peran, Ludovic and Giang, Minh and Farabet, Cl{\'e}ment and Vinyals, Oriol and Dean, Jeff and Kavukcuoglu, Koray and Hassabis, Demis and Ghahramani, Zoubin and Eck, Douglas},
  institution  = {Google DeepMind},
  year         = {2024},
  url          = {https://arxiv.org/abs/2403.08295}
}

@inproceedings{subramanian-etal-2020-medicat,
    title = "{M}ed{IC}a{T}: A Dataset of Medical Images, Captions, and Textual References",
    author = "Subramanian, Sanjay  and
      Wang, Lucy Lu  and
      Bogin, Ben  and
      Mehta, Sachin  and
      van Zuylen, Madeleine  and
      Parasa, Sravanthi  and
      Singh, Sameer  and
      Gardner, Matt  and
      Hajishirzi, Hannaneh",
    editor = "Cohn, Trevor  and
      He, Yulan  and
      Liu, Yang",
    booktitle = "Findings of the Association for Computational Linguistics: EMNLP 2020",
    month = nov,
    year = "2020",
    address = "Online",
    publisher = "Association for Computational Linguistics",
    url = "https://aclanthology.org/2020.findings-emnlp.191/",
    doi = "10.18653/v1/2020.findings-emnlp.191",
    pages = "2112--2120"
}

@inproceedings{wu-etal-2024-scimmir,
    title = "{S}ci{MMIR}: Benchmarking Scientific Multi-modal Information Retrieval",
    author = "Wu, Siwei  and
      Li, Yizhi  and
      Zhu, Kang  and
      Zhang, Ge  and
      Liang, Yiming  and
      Ma, Kaijing  and
      Xiao, Chenghao  and
      Zhang, Haoran  and
      Yang, Bohao  and
      Chen, Wenhu  and
      Huang, Wenhao  and
      Al Moubayed, Noura  and
      Fu, Jie  and
      Lin, Chenghua",
    editor = "Ku, Lun-Wei  and
      Martins, Andre  and
      Srikumar, Vivek",
    booktitle = "Findings of the Association for Computational Linguistics: ACL 2024",
    month = aug,
    year = "2024",
    address = "Bangkok, Thailand",
    publisher = "Association for Computational Linguistics",
    url = "https://aclanthology.org/2024.findings-acl.746/",
    doi = "10.18653/v1/2024.findings-acl.746",
    pages = "12560--12574"}

@article{chen2024expanding,
  title={Expanding Performance Boundaries of Open-Source Multimodal Models with Model, Data, and Test-Time Scaling},
  author={Chen, Zhe and Wang, Weiyun and Cao, Yue and Liu, Yangzhou and Gao, Zhangwei and Cui, Erfei and Zhu, Jinguo and Ye, Shenglong and Tian, Hao and Liu, Zhaoyang and others},
  journal={arXiv preprint arXiv:2412.05271},
  year={2024}
}

@article{wang2024mpo,
  title={Enhancing the Reasoning Ability of Multimodal Large Language Models via Mixed Preference Optimization},
  author={Wang, Weiyun and Chen, Zhe and Wang, Wenhai and Cao, Yue and Liu, Yangzhou and Gao, Zhangwei and Zhu, Jinguo and Zhu, Xizhou and Lu, Lewei and Qiao, Yu and Dai, Jifeng},
  journal={arXiv preprint arXiv:2411.10442},
  year={2024}
}

@article{chen2024far,
  title={How Far Are We to GPT-4V? Closing the Gap to Commercial Multimodal Models with Open-Source Suites},
  author={Chen, Zhe and Wang, Weiyun and Tian, Hao and Ye, Shenglong and Gao, Zhangwei and Cui, Erfei and Tong, Wenwen and Hu, Kongzhi and Luo, Jiapeng and Ma, Zheng and others},
  journal={arXiv preprint arXiv:2404.16821},
  year={2024}
}

@inproceedings{chen2024internvl,
  title={Internvl: Scaling up vision foundation models and aligning for generic visual-linguistic tasks},
  author={Chen, Zhe and Wu, Jiannan and Wang, Wenhai and Su, Weijie and Chen, Guo and Xing, Sen and Zhong, Muyan and Zhang, Qinglong and Zhu, Xizhou and Lu, Lewei and others},
  booktitle={Proceedings of the IEEE/CVF Conference on Computer Vision and Pattern Recognition},
  pages={24185--24198},
  year={2024}
}

@misc{mistral2025small31,
  title        = {Mistral-Small-3.1-24B-Instruct-2503},
  author       = {{Mistral AI}},
  year         = {2025},
  howpublished = {\url{https://huggingface.co/mistralai/Mistral-Small-3.1-24B-Instruct-2503}},
  note         = {Apache 2.0 License}
}

@techreport{phi4mini2025,
  title        = {Phi-4-Mini Technical Report: Compact yet Powerful Multimodal Language Models via Mixture-of-LoRAs},
  author       = {Abouelenin, Abdelrahman and Ashfaq, Atabak and Atkinson, Adam and Awadalla, Hany and Bach, Nguyen and Bao, Jianmin and Benhaim, Alon and Cai, Martin and Chaudhary, Vishrav and Chen, Congcong and Chen, Dong and Chen, Dongdong and Chen, Junkun and Chen, Weizhu and Chen, Yen-Chun and Chen, Yi-ling and Dai, Qi and Dai, Xiyang and Fan, Ruchao and Gao, Mei and Gao, Min and Garg, Amit and Goswami, Abhishek and Hao, Junheng and Hendy, Amr and Hu, Yuxuan and Jin, Xin and Khademi, Mahmoud and Kim, Dongwoo and Kim, Young Jin and Lee, Gina and Li, Jinyu and Li, Yunsheng and Liang, Chen and Lin, Xihui and Lin, Zeqi and Liu, Mengchen and Liu, Yang and Lopez, Gilsinia and Luo, Chong and Madan, Piyush and Mazalov, Vadim and Mitra, Arindam and Mousavi, Ali and Nguyen, Anh and Pan, Jing and Perez-Becker, Daniel and Platin, Jacob and Portet, Thomas and Qiu, Kai and Ren, Bo and Ren, Liliang and Roy, Sambuddha and Shang, Ning and Shen, Yelong and Singhal, Saksham and Som, Subhojit and Song, Xia and Sych, Tetyana and Vaddamanu, Praneetha and Wang, Shuohang and Wang, Yiming and Wang, Zhenghao and Wu, Haibin and Xu, Haoran and Xu, Weijian and Yang, Yifan and Yang, Ziyi and Yu, Donghan and Zabir, Ishmam and Zhang, Jianwen and Zhang, Li Lyna and Zhang, Yunan and Zhou, Xiren},
  institution  = {Microsoft Research},
  year         = {2025},
  archivePrefix= {arXiv},
  primaryClass = {cs.LG}
}

@techreport{pixtral12b2024,
  title        = {Pixtral 12B},
  author       = {Agrawal, Pravesh and Antoniak, Szymon and Bou Hanna, Emma and Bout, Baptiste and Chaplot, Devendra and Chudnovsky, Jessica and Costa, Diogo and De Monicault, Baudouin and Garg, Saurabh and Gervet, Theophile and Ghosh, Soham and Héliou, Amélie and Jacob, Paul and Jiang, Albert Q. and Khandelwal, Kartik and Lacroix, Timothée and Lample, Guillaume and Las Casas, Diego and Lavril, Thibaut and Le Scao, Teven and Lo, Andy and Marshall, William and Martin, Louis and Mensch, Arthur and Muddireddy, Pavankumar and Nemychnikova, Valera and Pellat, Marie and Von Platen, Patrick and Raghuraman, Nikhil and Rozière, Baptiste and Sablayrolles, Alexandre and Saulnier, Lucile and Sauvestre, Romain and Shang, Wendy and Soletskyi, Roman and Stewart, Lawrence and Stock, Pierre and Studnia, Joachim and Subramanian, Sandeep and Vaze, Sagar and Wang, Thomas and Yang, Sophia},
  institution  = {Mistral AI},
  year         = {2024},
  url          = {https://mistral.ai/news/pixtral-12b}
}

@techreport{qwen25vl2024,
  title        = {Qwen2.5-VL Technical Report},
  author       = {Bai, Shuai and Chen, Keqin and Liu, Xuejing and Wang, Jialin and Ge, Wenbin and Song, Sibo and Dang, Kai and Wang, Peng and Wang, Shijie and Tang, Jun and Zhong, Humen and Zhu, Yuanzhi and Yang, Mingkun and Li, Zhaohai and Wan, Jianqiang and Wang, Pengfei and Ding, Wei and Fu, Zheren and Xu, Yiheng and Ye, Jiabo and Zhang, Xi and Xie, Tianbao and Cheng, Zesen and Zhang, Hang and Yang, Zhibo and Xu, Haiyang and Lin, Junyang},
  institution  = {Alibaba Group},
  year         = {2024},
  url          = {https://arxiv.org/abs/2406.16855},
}

@techreport{meta2024llama3,
  title        = {The Llama 3 Herd of Models},
  author       = {{Meta AI}},
  institution  = {Meta},
  year         = {2024},
  url          = {https://ai.meta.com/blog/llama-3}
}

@misc{openai2025gpt5,
  title        = {Introducing GPT-5 for Developers},
  author       = {{OpenAI}},
  year         = {2025},
  howpublished = {\url{https://openai.com/index/introducing-gpt-5-for-developers/}},
  month        = aug
}

@misc{openai2024gpt4o,
  title        = {Hello GPT-4o},
  author       = {{OpenAI}},
  year         = {2024},
  howpublished = {\url{https://openai.com/index/hello-gpt-4o/}}
}

@inproceedings{vladika-matthes-2023-scientific,
    title = "Scientific Fact-Checking: A Survey of Resources and Approaches",
    author = "Vladika, Juraj  and
      Matthes, Florian",
    editor = "Rogers, Anna  and
      Boyd-Graber, Jordan  and
      Okazaki, Naoaki",
    booktitle = "Findings of the Association for Computational Linguistics: ACL 2023",
    month = jul,
    year = "2023",
    address = "Toronto, Canada",
    publisher = "Association for Computational Linguistics",
    url = "https://aclanthology.org/2023.findings-acl.387/",
    doi = "10.18653/v1/2023.findings-acl.387",
    pages = "6215--6230"
}

@inproceedings{PubHealth2020,
  author    = {Kotonya, Neema and Toni, Francesca},
  title     = {Explainable Automated Fact-Checking for Public Health Claims},
  booktitle = {Proceedings of the 2020 Conference on Empirical Methods in Natural Language Processing (EMNLP)},
  pages     = {7740--7754},
  year      = {2020},
  address   = {Online},
  publisher = {Association for Computational Linguistics}
}

@inproceedings{COVIDFact2021,
  author    = {Saakyan, Arkadiy and Chakrabarty, Tuhin and Muresan, Smaranda},
  title     = {COVID-Fact: Fact Extraction and Verification of Real-World Claims on the COVID-19 Pandemic},
  booktitle = {Proceedings of the 59th Annual Meeting of the Association for Computational Linguistics and the 11th International Joint Conference on Natural Language Processing (Volume 1: Long Papers)},
  pages     = {2116--2129},
  year      = {2021},
  address   = {Online},
  publisher = {Association for Computational Linguistics}
}

@inproceedings{CoVERT2022,
  author    = {Mohr, Isabelle and W{\"u}hrl, Amelie and Klinger, Roman},
  title     = {CoVERT: A Corpus of Fact-Checked Biomedical COVID-19 Tweets},
  booktitle = {Proceedings of the Language Resources and Evaluation Conference (LREC)},
  pages     = {244--257},
  year      = {2022},
  address   = {Marseille, France},
  publisher = {European Language Resources Association}
}

@inproceedings{MOCHEG,
author = {Yao, Barry Menglong and Shah, Aditya and Sun, Lichao and Cho, Jin-Hee and Huang, Lifu},
title = {End-to-End Multimodal Fact-Checking and Explanation Generation: A Challenging Dataset and Models},
year = {2023},
isbn = {9781450394086},
publisher = {Association for Computing Machinery},
address = {New York, NY, USA},
url = {https://doi.org/10.1145/3539618.3591879},
doi = {10.1145/3539618.3591879},
booktitle = {Proceedings of the 46th International ACM SIGIR Conference on Research and Development in Information Retrieval},
pages = {2733–2743},
numpages = {11},
location = {Taipei, Taiwan},
series = {SIGIR '23}
}

@article{zou2026diagpaper,
  title={DIAGPaper: Diagnosing Valid and Specific Weaknesses in Scientific Papers via Multi-Agent Reasoning},
  author={Zou, Zhuoyang and Ansari, Abolfazl and Zhang, Delvin Ce and Lee, Dongwon and Yin, Wenpeng},
  journal={arXiv preprint arXiv:2601.07611},
  year={2026}
}

@inproceedings{cikm_Tahmasebi24,
    author = {Tahmasebi, Sahar and M{\"u}ller-Budack, Eric and Ewerth, Ralph},
    title = {Multimodal Misinformation Detection using Large Vision-Language Models},
    booktitle = {Proceedings of the ACM International Conference on Information and Knowledge Management (CIKM)},
    year = {2024}
}

@misc{chen2024huatuogptvisioninjectingmedicalvisual,
      title={HuatuoGPT-Vision, Towards Injecting Medical Visual Knowledge into Multimodal LLMs at Scale}, 
      author={Junying Chen and Ruyi Ouyang and Anningzhe Gao and Shunian Chen and Guiming Hardy Chen and Xidong Wang and Ruifei Zhang and Zhenyang Cai and Ke Ji and Guangjun Yu and Xiang Wan and Benyou Wang},
      year={2024},
      eprint={2406.19280},
      archivePrefix={arXiv},
      primaryClass={cs.CV},
      url={https://arxiv.org/abs/2406.19280}, 
}

@misc{diggelmann2020climatefever,
      title={CLIMATE-FEVER: A Dataset for Verification of Real-World Climate Claims},
      author={Thomas Diggelmann and Jordan Boyd-Graber and Jannis Bulian and Massimiliano Ciaramita and Markus Leippold},
      year={2020},
      eprint={2012.00614},
      archivePrefix={arXiv},
      primaryClass={cs.CL}
}

@inproceedings{wang2025piecing,
  title={Piecing it all together: Verifying multi-hop multimodal claims},
  author={Wang, Haoran and Rangapur, Aman and Xu, Xiongxiao and Liang, Yueqing and Gharwi, Haroon and Yang, Carl and Shu, Kai},
  booktitle={Proceedings of the 31st International Conference on Computational Linguistics},
  pages={7453--7469},
  year={2025}
}

@article{dmonte2024claim,
  title={Claim verification in the age of large language models: A survey},
  author={Dmonte, Alphaeus and Oruche, Roland and Zampieri, Marcos and Calyam, Prasad and Augenstein, Isabelle},
  journal={arXiv preprint arXiv:2408.14317},
  year={2024}
}

@inproceedings{thorne-etal-2018-fever,
    title = "{FEVER}: a Large-scale Dataset for Fact Extraction and {VER}ification",
    author = "Thorne, James  and
      Vlachos, Andreas  and
      Christodoulopoulos, Christos  and
      Mittal, Arpit",
    editor = "Walker, Marilyn  and
      Ji, Heng  and
      Stent, Amanda",
    booktitle = "Proceedings of the 2018 Conference of the North {A}merican Chapter of the Association for Computational Linguistics: Human Language Technologies, Volume 1 (Long Papers)",
    month = jun,
    year = "2018",
    address = "New Orleans, Louisiana",
    publisher = "Association for Computational Linguistics",
    url = "https://aclanthology.org/N18-1074/",
    doi = "10.18653/v1/N18-1074",
    pages = "809--819"
}

@article{hu2022lora,
  title={Lora: Low-rank adaptation of large language models.},
  author={Hu, Edward J and Shen, Yelong and Wallis, Phillip and Allen-Zhu, Zeyuan and Li, Yuanzhi and Wang, Shean and Wang, Liang and Chen, Weizhu and others},
  journal={Iclr},
  volume={1},
  number={2},
  pages={3},
  year={2022}
}

@article{ethicsofimage,
author = {Skulmowski, Alexander and Engel-Hermann, Patricia},
title = {The ethics of erroneous AI-generated scientific figures},
year = {2025},
issue_date = {Apr 2025},
publisher = {Kluwer Academic Publishers},
address = {USA},
volume = {27},
number = {2},
issn = {1388-1957},
url = {https://doi.org/10.1007/s10676-025-09835-4},
doi = {10.1007/s10676-025-09835-4},
journal = {Ethics and Inf. Technol.},
month = jun,
numpages = {10}
}

@misc{ramesh2021zeroshottexttoimagegeneration,
      title={Zero-Shot Text-to-Image Generation}, 
      author={Aditya Ramesh and Mikhail Pavlov and Gabriel Goh and Scott Gray and Chelsea Voss and Alec Radford and Mark Chen and Ilya Sutskever},
      year={2021},
      eprint={2102.12092},
      archivePrefix={arXiv},
      primaryClass={cs.CV},
      url={https://arxiv.org/abs/2102.12092}, 
}

@misc{rombach2022highresolutionimagesynthesislatent,
      title={High-Resolution Image Synthesis with Latent Diffusion Models}, 
      author={Robin Rombach and Andreas Blattmann and Dominik Lorenz and Patrick Esser and Björn Ommer},
      year={2022},
      eprint={2112.10752},
      archivePrefix={arXiv},
      primaryClass={cs.CV},
      url={https://arxiv.org/abs/2112.10752}, 
}
\bibliographystyle{colm2026_conference}

\appendix
\appendix
\label{sec:appendix}

\section*{Limitations}
While {\name} advances multimodal verification, we acknowledge certain limitations. First, despite our rigorous three-stage expert evaluation, fully human-verifying all 469K instances is infeasible and labor-intensive. Consequently, subtle generation biases may persist, though our expert evaluation and statistical analysis indicate they do not negatively impact reliability and model learnability. Second, {\name} focuses exclusively on static scientific figures, leaving dynamic visual components like videos or interactive 3D structures to future work. Finally, our dataset is exclusively in English. Future research should prioritize expanding language coverage and integrating diverse evidence modalities.

\section{{\name} Configuration and Reproducibility}
\subsection{Claim-Indicating Verbs}
\label{sec:claim_indicating}

To filter for explicit conclusions in the {\name} dataset, we utilized the following list of claim-indicating base verbs and their morphological variations: show, shows, showed, demonstrate, demonstrates, demonstrated, depict, depicts, illustrate, illustrates, present, presents, display, displays, indicate, indicates.

\subsection{Prompt Templates for {\name}}
\label{sec:appendix_prompts_generation}

In this section, we provide the exact prompt templates used to instruct our large language models for both the claim perturbation and explanation generation tasks. 

\vspace{2mm}

\begin{tcolorbox}[
    colback=black!3,
    colframe=black!20,
    boxrule=0.5pt,
    arc=4pt,
    left=8pt,
    right=8pt,
    top=7pt,
    bottom=7pt
]
\textbf{Medical Claim Perturbation Prompt.}

\textbf{Image:} \{image\} \\
\textbf{Label:} \{Support\} \\
\textbf{Claim:} \{claim\} \\
\textbf{Caption:} \{caption\} \\
\textbf{Perturbation Taxonomy:} \{Available Perturbations\}

\vspace{3pt}
You are a medical expert tasked with generating a refuted version of the given medical claim. Your goal is to produce a clear, strong, and scientifically sound opposite claim by applying the most suitable perturbation from the provided taxonomy. 

The generated claim must be entirely opposite in meaning and medical implication. Do not simply negate the original claim using words like `not', `no', or `without'. Instead, generate a completely new claim that semantically contradicts the original.

The applied perturbation must change the veracity of the claim, rendering it incorrect in light of the original evidence. Maintain a neutral, professional, and medically appropriate tone. Choose the best-matching perturbation category from the taxonomy below to construct a concise, direct, and well-phrased refuted claim of similar length to the original.

\textbf{Output format:} \\
Perturbation: $\langle$best-matching perturbation category that changes the veracity label of the claim$\rangle$ \\
Refuted Claim: $\langle$refuted claim$\rangle$

If none of the listed categories produce a valid contradiction, return empty as perturbation taxonomy.
\end{tcolorbox}

\begin{tcolorbox}[
    colback=black!3,
    colframe=black!20,
    boxrule=0.5pt,
    arc=4pt,
    left=8pt,
    right=8pt,
    top=7pt,
    bottom=7pt
]
\textbf{Explanation Generation Prompt.}

\textbf{Image:} \{image\} \\
\textbf{Label:} \{label\} \\
\textbf{Claim:} \{claim\} \\
\textbf{Caption:} \{caption\}

You are a scientific expert. Based solely on the provided image and caption, briefly explain why the given label is appropriate for this scientific claim. 

Rationalize your decision using evidence grounded exclusively in the figure and caption to justify the label. Keep your explanation to approximately 100 words and do not use any external sources of evidence.
\end{tcolorbox}

\subsection{LLM-as-a-Judge Evaluation}
\label{sec:llm_judge_score}

To ensure high data precision and alignment, we implemented an automated filtering pipeline. The system evaluates the alignment between the claim and the evidence set (figure and caption) on a granular 1--10 scale using the following prompt:

\begin{tcolorbox}[
    colback=black!3,
    colframe=black!20,
    boxrule=0.5pt,
    arc=4pt,
    left=8pt,
    right=8pt,
    top=7pt,
    bottom=7pt
]
\textbf{Evidence Grounding Evaluation Prompt}

\textbf{Image} \{image\} \\
\textbf{Label} \{label\} \\
\textbf{Claim} \{claim\} \\
\textbf{Caption} \{caption\}

\vspace{3pt}
You are an expert tasked with evaluating how well the provided evidence supports or refutes a given scientific claim

\textbf{Task}
Based on the claim and the available visual and textual evidence rate how well the evidence justifies the \{expected label\} label on a scale of 1 to 10

Scoring Guide
\begin{itemize}[noitemsep,topsep=0pt,leftmargin=*]
    \item 10 Perfect
    \item 1 Very Poor
\end{itemize}

Instructions
\begin{itemize}[noitemsep,topsep=0pt,leftmargin=*]
    \item Step 1 Consider the relevance completeness and clarity of the evidence
    \item Step 2 Rate how well the evidence aligns with the claim strictly from 1 to 10
    \item Step 3 Provide a short justification for your rating
\end{itemize}

\textbf{Output Format}
Score \{number from 1 to 10\} \\
Justification \{brief explanation\}
\end{tcolorbox}

\subsection{Threshold Tuning and Filtering}
To define the rejection threshold, we analyzed the distribution of grounding scores alongside a manual quality evaluation. We observed that instances receiving scores of 4 or below consistently corresponded to insufficient visual evidence (NEI), whereas scores of 5 and above provided actionable multimodal grounding. We further found that modest variations around this threshold do not materially affect the filtering outcomes, indicating the robustness of this choice. Applying this strict threshold, the filtering protocol systematically discarded low-confidence generations, ensuring that the final benchmark contains only high-fidelity, visually grounded examples.


\section{Expert Evaluation Protocols}
\label{sec:human_eval}

We detail our evaluation protocols, recruitment criteria, and instructions provided to our human evaluators in this section. Our evaluation was split into distinct phases for claim verification and explanation faithfulness. These phases were strictly conducted across separate human intelligent tasks (HIT) with entirely different groups of annotators to prevent cognitive bias and label leakage.

\subsection{Recruitment Demographics and Compensation}
We recruited 92 domain experts via the Prolific platform. To ensure data reliability participation was restricted to residents of the United States with substantial prior platform experience. On average annotators had approximately 2524 previously approved Prolific submissions indicating high familiarity with rigorous annotation tasks.

We utilized Prolific pre-screening features to strictly align evaluator expertise with the scientific domain of each dataset subset
\begin{itemize}[noitemsep,topsep=0pt]
    \item \textbf{Medical Domain Evaluators} Participants holding graduate or doctoral degrees in Medicine Nursing or Health Sciences
    \item \textbf{General Science Evaluators} Participants holding graduate degrees in Science Technology Engineering or Mathematics
\end{itemize}

Participants were compensated at an average rate of \$14.00/hour exceeding standard platform minimums for scientific reasoning tasks. All study protocols and data collection procedures were approved by our Institutional Review Board (IRB).

\subsection{Inter Annotator Agreement Evaluation}
The first evaluation phase measured whether human experts could truthfully predict the verification label relying strictly on the provided claim and scientific evidence.

\noindent\textbf{Task Setup and Input} Annotators were presented with the scientific figure the caption and the claim. Annotators were strictly not shown the ground truth label or any model generated explanations during this phase. This setting ensured that their judgment and agreement on cognitive task relied only on the visual and textual evidence. Each instance in a set of 200 claims was assigned to two independent domain experts to rigorously measure IAA agreement.An instance representation of the task interface presented to the annotators is illustrated in the example block below.

\begin{tcolorbox}[
    colback=black!3,
    colframe=black!20,
    boxrule=0.5pt,
    arc=4pt,
    left=8pt,
    right=8pt,
    top=7pt,
    bottom=7pt
]
\textbf{Claim Verification Task Example}

\vspace{2mm}
\begin{center}
\includegraphics[width=0.6\linewidth]{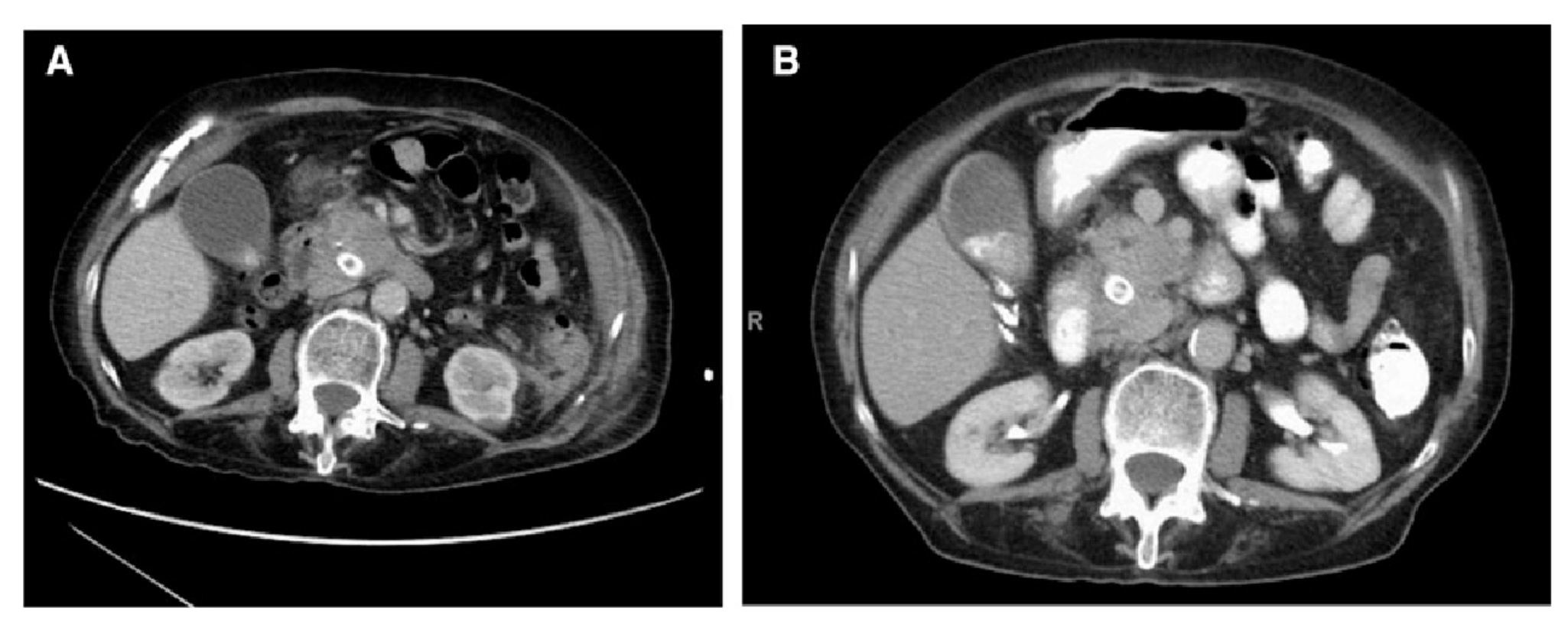}
\label{fig:instruction}
\end{center}
\vspace{2mm}

\textbf{Caption} FIG 1 A February 2015 CT scan B October 2015 CT scan

\textbf{Claim 1} A follow up CT scan performed 8 months after a single dose of chemotherapy demonstrates stable disease compared to the prior scan

\textbf{Question} Is this claim supported or refuted by the image and caption:

\begin{itemize}[noitemsep,topsep=0pt,leftmargin=*]

    \item This claim is supported by evidence

    \item This claim is refuted by evidence

    \item I cannot verify this specific example

\end{itemize}

\end{tcolorbox}

\subsection{Annotator Alignment Evaluation}
The second cognitive task evaluated the large scale alignment of the generated claims with their assigned verification labels.

\noindent\textbf{Task Setup and Input} Each instance in the broader dataset was assigned to a single domain expert to maximize coverage at scale. A set of 1,295 claims was evaluated by domain experts to establish a standard gold audit set. We then used this gold audit set to validate the reliability of this single expert approach through a statistical gold set audit. When comparing the full dataset results to a human verified gold set we observed high ranking stability with a Spearman correlation of 0.943 for the general science domain and 0.899 for the medical domain. For this phase we report the agreement metric as the percentage of instances where the expert explicitly verified that the claim was fully consistent with the provided evidence.

\subsection{Explanation Quality Evaluation}
To evaluate the scientific quality of 1,096 generated rationales we conducted a third evaluation phase utilizing a completely different set of domain experts. 

\noindent\textbf{Task Setup and Input} We provided these experts with the visual evidence the caption the scientific claim the ground truth label and explanation. We then asked the annotators to critically evaluate whether the explanation truthfully justified the label and properly aligned with the scientific evidence without hallucinating facts or numbers.

\section{arXiv Subject Categories}
\label{sec:arxiv_categories}

Table~\ref{tab:arxiv_categories_table} details the 16 specific arXiv subject categories and their corresponding short-form codes used to construct the general science split of the {\name} dataset.

\begin{table}[t]

\centering

\small

\setlength{\tabcolsep}{3pt}

\begin{tabular}{ll}

\hline

\textbf{Short Form} & \textbf{Full Name} \\

\hline

astro     & Astrophysics \\

cond      & Condensed Matter Physics \\

cs        & Computer Science \\

econ      & Economics \\

eess      & Electrical Engineering and Systems Science \\

grqc      & General Relativity and Quantum Cosmology \\

hep\_ex   & High Energy Physics\\





math      & Mathematics \\

math\_ph  & Mathematical Physics \\

nlin      & Nonlinear Sciences \\

nucl\_th  & Nuclear Theory \\

physics   & Physics (General) \\

qbio      & Quantitative Biology \\

qfin      & Quantitative Finance \\

quant\_ph & Quantum Physics \\

stat      & Statistics \\
\hline
\end{tabular}
\caption{arXiv subject categories: short-form codes and full names.}
\label{tab:arxiv_categories_table}

\end{table}

\section{Baseline Evaluation Prompts}
\label{appendix:baseline_prompts}

The following zero-shot prompt templates were used to evaluate all baseline vision-language models on the {\name} test sets. To ensure a fair comparison, the same prompts were applied across all open-source and closed-source models.

\begin{tcolorbox}[
    colback=black!3,
    colframe=black!20,
    boxrule=0.5pt,
    arc=4pt,
    left=8pt,
    right=8pt,
    top=7pt,
    bottom=7pt
]
\textbf{Claim Verification Prompt (Multimodal Baseline).}

\textbf{Image:} \{image\} \\
\textbf{Claim:} \{claim\} \\
\textbf{Caption:} \{caption\}

You are a scientific expert. Based only on the provided image and caption, decide whether the caption supports the claim or refutes it. \\
Reply with a single word only: yes or no.
\end{tcolorbox}

\begin{tcolorbox}[
    colback=black!3,
    colframe=black!20,
    boxrule=0.5pt,
    arc=4pt,
    left=8pt,
    right=8pt,
    top=7pt,
    bottom=7pt
]
\textbf{Explanation Generation Prompt (Baseline).}

\textbf{Image:} \{image\} \\
\textbf{Label:} \{label\} \\
\textbf{Claim:} \{claim\} \\
\textbf{Caption:} \{caption\}

You are a scientific expert. Based only on the provided image and caption, briefly explain why the given label is appropriate for this scientific claim.

Keep your explanation around 100 words and only limited to scope of evidence.
\end{tcolorbox}

\section{Explanation Evaluation Details}
\label{sec:appendix_explanation_eval}

\subsection{LLM as a Judge Evaluation Metrics and Prompt}
\label{sec:appendix_llm_judge}

To rigorously evaluate the semantic quality of the generated explanations we prompted DeepSeek-3.2 to act as a scientific judge. The judge evaluated each generated explanation across five distinct semantic dimensions on a scale of 1 to 10. 

The five evaluated dimensions are defined below

\begin{itemize}[noitemsep,topsep=0pt]
    \item \textbf{Correctness} Assesses if the generated explanation contains accurate scientific facts without hallucinated details.
    \item \textbf{Completeness} Checks if the explanation covers all the necessary reasoning steps to fully justify the claim.
    \item \textbf{Entailment} Evaluates whether the explanation logically follows from the provided visual and textual evidence.
    \item \textbf{Relevance} Measures if the explanation stays strictly on topic without introducing outside or tangential information.
    \item \textbf{Clarity} Determines if the text is logically structured and easy for a scientific reader to understand.
\end{itemize}

\vspace{2mm}
The exact prompt utilized for this semantic evaluation is provided below.

\begin{tcolorbox}[
    colback=black!3,
    colframe=black!20,
    boxrule=0.5pt,
    arc=4pt,
    left=8pt,
    right=8pt,
    top=7pt,
    bottom=7pt
]
\textbf{Explanation Quality Evaluation Prompt.}

\textbf{Image:} \{image\} \\
\textbf{Caption:} \{caption\} \\
\textbf{Claim:} \{claim\} \\
\textbf{Model Output Explanation:} \{generated\_explanation\}

\vspace{3pt}
You are a scientific reasoning expert. Evaluate the quality of the Generated Explanation based on the provided visual evidence, caption, and claim. Rate the text on a 1--10 scale (higher is better) across five dimensions: entailment, relevance, correctness, completeness, and clarity. 

\textbf{Output Format:} \\
\{ "entailment" $\langle$1-10$\rangle$, "relevance" $\langle$1-10$\rangle$, "correctness" $\langle$1-10$\rangle$, "completeness" $\langle$1-10$\rangle$, "clarity" $\langle$1-10$\rangle$ \}
\end{tcolorbox}

\subsection{Case Study in Reasoning and Linguistic Artifacts}
\label{sec:appendix_case_study_artifacts}

We present a qualitative case study highlighting failure modes in model-generated explanations. Even when the predicted label is correct, the model produces unexpected linguistic artifacts, indicating a gap between decision accuracy and faithful explanation.

\begin{tcolorbox}[
    colback=black!3,
    colframe=black!20,
    boxrule=0.5pt,
    arc=4pt,
    left=8pt,
    right=8pt,
    top=7pt,
    bottom=7pt
]
\textbf{Qualitative Case Study Qwen2.5 VL 7B}

\vspace{2mm}
\begin{center}
\includegraphics[width=0.4\linewidth]{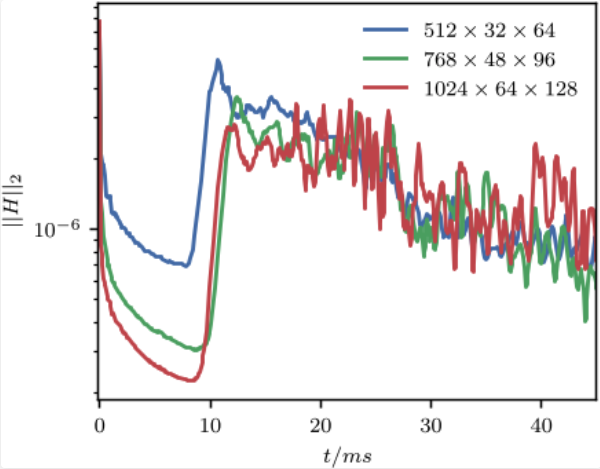}
\end{center}
\vspace{2mm}

\textbf{Caption} FIG 13 The L2 norm of the Hamiltonian constraint for the bar mode unstable star. Prior to the onset of instability we see convergence to zero at 1.5 order. 

\textbf{Claim} The L2 norm of the Hamiltonian constraint decreases across resolutions from a value above $10^{-6}$ at $t/ms = 0$ to a value below $10^{-6}$ at $t/ms = 10$.

\textbf{Qwen2.5 VL 7B Output} The label support is appropriate because the figure shows convergence to zero of the L2 norm which indicates stability. Despite a subsequent instability \begin{CJK*}{UTF8}{gbsn}引起的\end{CJK*} Constraint violations remain manageable and do not exhibit exponential increase consistent with the scientific claim.
\end{tcolorbox}

\vspace{2mm}
The model fails on Correctness by ignoring the specific numerical bounds of the claim and instead relying on a general heuristic of convergence to zero. Furthermore the unprompted insertion of the Chinese characters \begin{CJK*}{UTF8}{gbsn}引起的\end{CJK*} into the technical English rationale reveals a breakdown in the decoding pipeline. These qualitative artifacts confirm that high verification accuracy does not guarantee either precise scientific grounding or linguistic reliability in multilingual vision language models.

\end{document}